\documentclass{article}

\usepackage{microtype}
\usepackage{graphicx}
\usepackage{subcaption}
\usepackage{booktabs} 

\usepackage{hyperref}

\usepackage[accepted]{icml2024}

\usepackage{amsmath}
\usepackage{amssymb}
\usepackage{mathtools}
\usepackage{amsthm}

\usepackage[capitalize,noabbrev]{cleveref}

\usepackage{algorithm}
\usepackage{algorithmic}
\usepackage[frozencache,cachedir=minted-cache]{minted}
\setminted{breaklines}
\usepackage{amsmath}
\usepackage{multirow}

\theoremstyle{plain}

\theoremstyle{definition}

\theoremstyle{remark}

\usepackage[textsize=tiny]{todonotes}

\icmltitlerunning{Rethinking Momentum Knowledge Distillation in Online Continual Learning}

\begin{document}

\twocolumn[
\icmltitle{Rethinking Momentum Knowledge Distillation in Online Continual Learning}



\icmlsetsymbol{equal}{*}

\begin{icmlauthorlist}
\icmlauthor{Nicolas Michel}{paris}
\icmlauthor{Maorong Wang}{tokyo}
\icmlauthor{Ling Xiao}{tokyo}
\icmlauthor{Toshihiko Yamasaki}{tokyo}
\end{icmlauthorlist}

\icmlaffiliation{paris}{Univ Gustave Eiffel, CNRS, LIGM, F-77454 Marne-la-Vallée, France}
\icmlaffiliation{tokyo}{The University of Tokyo, 7-3-1 Hongo, Bunkyo-ku, Tokyo 113-8656, Japan}

\icmlcorrespondingauthor{Nicolas Michel}{nicolas.michel@esiee.fr}

\icmlkeywords{Machine Learning, ICML}

\vskip 0.3in
]



\printAffiliationsAndNotice{}  

\begin{abstract}
    Online Continual Learning (OCL) addresses the problem of training neural networks on a continuous data stream where multiple classification tasks emerge in sequence. In contrast to offline Continual Learning, data can be seen only once in OCL, which is a very severe constraint. In this context, replay-based strategies have achieved impressive results and most state-of-the-art approaches heavily depend on them. While Knowledge Distillation (KD) has been extensively used in offline Continual Learning, it remains under-exploited in OCL, despite its high potential. In this paper, we analyze the challenges in applying KD to OCL and give empirical justifications. We introduce a direct yet effective methodology for applying Momentum Knowledge Distillation (MKD) to many flagship OCL methods and demonstrate its capabilities to enhance existing approaches. In addition to improving existing state-of-the-art accuracy by more than $10\%$ points on ImageNet100, we shed light on MKD internal mechanics and impacts during training in OCL. We argue that similar to replay, MKD should be considered a central component of OCL. The code is available at \url{https://github.com/Nicolas1203/mkd_ocl}.
\end{abstract}

\section{Introduction}
\label{submission}

Over the past decade, Deep Neural Networks (DNNs) have demonstrated super-human performance in most vision tasks~\cite{he_deep_2015,redmon2016you,caron2021emerging,khosla_supervised_2020}. Nonetheless, current training procedures rely on strong assumptions. Specifically, during training, it is typically assumed that: 1) available data is independently and identically distributed (i.i.d.), and 2) all training data can be seen multiple times. Contrary to humans, DNNs are known to underperform or fail outright when these assumptions are not satisfied and suffer from Catastrophic Forgetting (CF)~\cite{french1999catastrophic,kirkpatrick_overcoming_2017}. Addressing these challenges, Online Continual Learning (OCL) explores methods to mitigate CF in scenarios that violate assumptions 1) and 2). This is done by learning from a continuous stream of \textbf{non-i.i.d. data} where \textbf{only one pass is allowed}. Formally, OCL considers a sequential learning setup with a sequence $\{\mathcal{T}_1,\cdots,\mathcal{T}_K\}$ of $K$ tasks, and $\mathcal{D}_k=(X_k, Y_k)$ the corresponding data-label pairs. For any value $k_1,k_2 \in \{1,\cdots,K\}$, if $k_1\neq k_2$ then $Y_{k_1}\cap Y_{k_2}=\emptyset$. This scenario is known to be especially difficult and numerous approaches have been proposed to address it~\cite{he_online_2021,guo_online_2022,mai_online_2021,mai_supervised_2021,caccia_new_2022,aljundi_online_2019,guo_dealing_2023, vedaldi_gdumb_2020, aljundi_gradient_2019, koh_online_2023, michel_learning_2023}. In this study, we focus on the Class Incremental Learning scenario \cite{hsu_re-evaluating_2019} for OCL.

\begin{figure}[t]
    \centering
    \includegraphics[width=\linewidth]{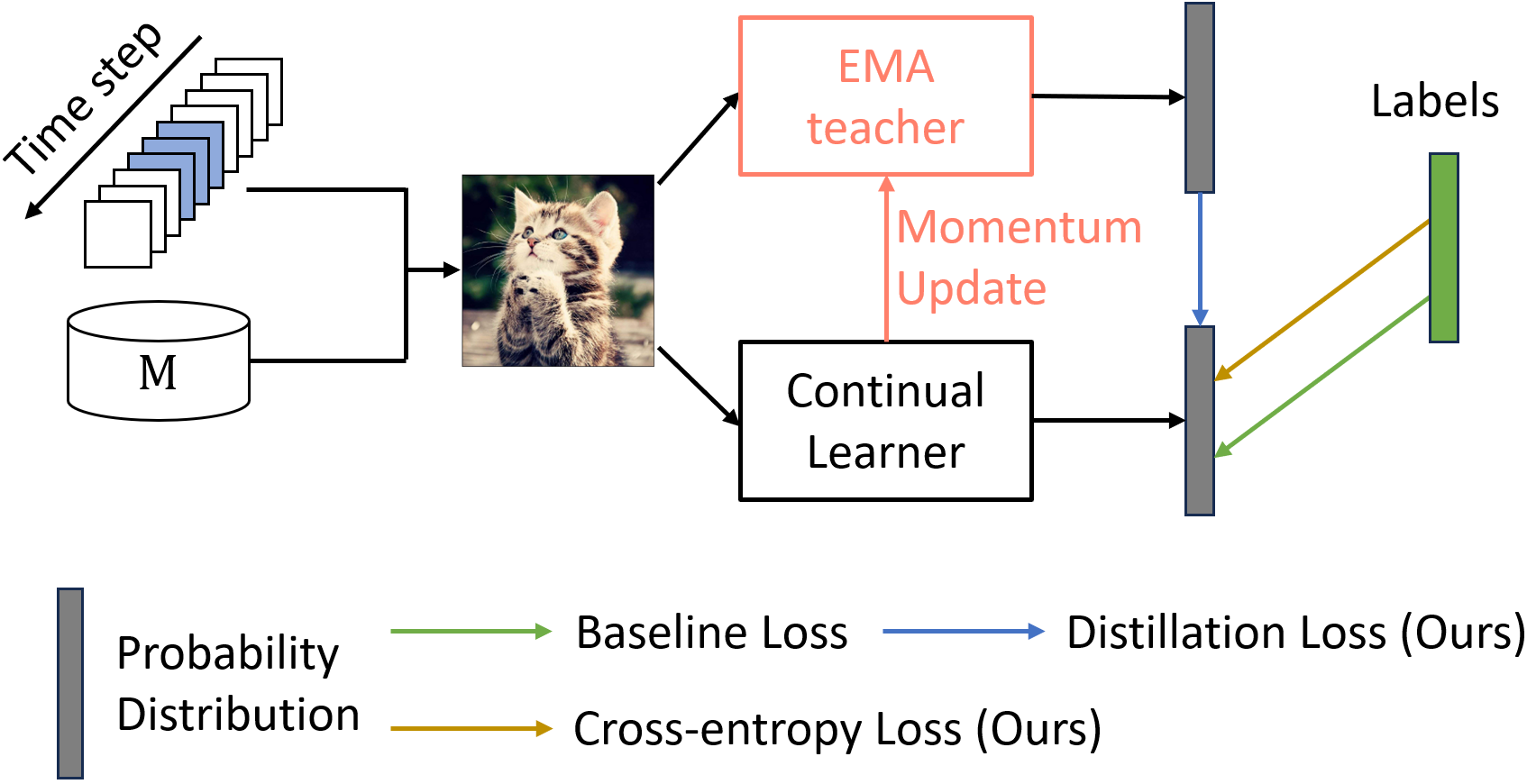}
    \caption{Overview of our MKD framework when applied to a baseline OCL method. Contrary to taking a snapshot at the end of each task, dynamic teacher address the key obstacles in OCL: teacher quality, teacher quantity, and unknown task boundaries.}
    \vspace{-.5em}
    \label{fig:teaser}
\end{figure}

Among various methods, Experience Replay (ER) approaches~\cite{rolnick_experience_2019,buzzega_dark_2020,khosla_supervised_2020,guo_online_2022,caccia_new_2022,michel_learning_2023,guo_dealing_2023} have demonstrated superior performances in OCL. The main component of this strategy is to store a small portion of previous samples to be used when training on new incoming samples. Current state-of-the-art methods in OCL mostly rely on combining replay strategies and specific loss designs. Unlike ER, only a few applications of Knowledge Distillation (KD) to OCL exist and present various limitations. DER~\cite{buzzega_dark_2020} stores previous sample logits and leverages knowledge distillation with ER but yields low performances. While MMKDDA~\cite{han_online_2022} tackles meta-learning with multi-level KD, it requires knowledge of total number of tasks and is computation intensive. Recently, SDP~\cite{koh_online_2023} proposes a hypo-exponential teacher for feature distillation in addition to ER. Even though SDP does not require task boundaries, it remains computationally expensive and architecture-dependent. In this work, we argue that KD has been rather overlooked by previous studies and can be efficiently adapted to OCL. Indeed, we believe that similarly to ER, KD plays an essential role in OCL and can be seamlessly combined with existing approaches.


Understanding the challenges specific to OCL is the key to explain why KD is not widely adopted in this context. Thus, we identify the three main KD challenges in OCL: Teacher Quality, Teacher Quantity and Unknown Task Boundaries. To overcome these challenges, we propose to take advantage of Momentum Knowledge Distillation (MKD)~\cite{caron2021emerging}. Although MKD is a straightforward strategy, our technical contribution is a procedure which allows us to seamlessly integrate MKD with existing state-of-the-art approaches and show considerable improvements, even when compared to other distillation methods. Additionally, we highlight that utilizing MKD for OCL addresses prominent OCL challenges such as task-recency bias~\cite{chrysakis2023online, mai_supervised_2021}, last layer bias~\cite{liang_new_2023,ahn_ss-il_2021,mai_supervised_2021, wu_large_2019}, feature drift~\cite{caccia_new_2022} and feature discrimination. In summary, the contributions of this paper are as follows:
\begin{itemize}
    \setlength\itemsep{0.1em}
    \item We identify the three main obstacles in applying KD to OCL and leverage MKD as a solution to overcome these challenges;
    \item We propose a strategy to seamlessly combine MKD with existing approaches and give insights on MKD internal mechanics and impacts during training in OCL;
    \item We experimentally demonstrate that MKD can significantly enhance the performance of existing methods.
\end{itemize}

\section{Related Work}

\subsection{KD in CL}
We review KD strategies in both offline and online CL. We define offline CL as the multi-epoch CL training.

\vspace{-1em}
\paragraph{KD in Offline CL}
Knowledge Distillation (KD)~\cite{hinton_distilling_2015} aims at transferring knowledge from a teacher model to a student model. This can be done by aligning their outputs, either in the logits space~\cite{hinton_distilling_2015,romero2014fitnets,zhao2022decoupled} or in the representation space~\cite{aguilar2020knowledge,tian_contrastive_2019}. There are numerous KD applications in offline CL~\cite{ahn_ss-il_2021,douillard2020podnet,rebuffi2017icarl,cha_co2l_2021,simon2021learning,hou2018lifelong,wang2022foster,pham2021dualnet}. A common practice is to save the model at the end of each task, treating it as a snapshot, and use this model as a teacher for distillation during subsequent task trainings~\cite{hou2018lifelong,cha_co2l_2021}. Given that each teacher has task-specific knowledge, SS-IL~\cite{ahn_ss-il_2021} leverages task-wise KD. There are also strategies that incorporate spatial distillation~\cite{douillard2020podnet} or feature compression~\cite{wang2022foster}.

\vspace{-1em}
\paragraph{KD in Online CL}
Although KD has been widely adopted in offline CL, its adoption in OCL remains limited. DER~\cite{buzzega_dark_2020} retains logits as well as data in memory for distillation in later stages. MMKDDA~\cite{han_online_2022} addresses meta-learning using multi-scale KD. Recently, SDP~\cite{koh_online_2023} introduced a teacher defined as a hypo-exponential moving average of current model for feature distillation. Nonetheless, these methods have their own constraints. DER exhibits suboptimal performance and scales poorly when increasing memory size; MMKDDA requires task boundaries and is resource-intensive; SDP is architecture dependent and computationally expensive.

\vspace{-0.5em}
\subsection{Blurry Task Boundaries}
A common assumption in CL is that task boundaries are distinctly recognized during training. Similar to the work of~\cite{michel_learning_2023}, we refer to this as \textit{clear} task boundaries. In OCL, however, we work on a continuous stream of incoming data, which makes \textit{clear} boundaries unrealistic.
In that sense, the concept of \textit{blurry} task boundary setting has emerged in recent studies~\cite{caccia_new_2022,michel_learning_2023,bang_online_2022}. The idea is to have a gradual transition between tasks with an intermediate stage where data from both tasks are available in the stream. In this study, we embrace the perspective of unknown task boundaries, referring to it as the \textit{blurry} setting, in opposition to the traditional \textit{clear} setting as in~\cite{michel_learning_2023}.

\vspace{-0.5em}
\subsection{Evaluation Metrics}
We use the accuracy averaged across all tasks after training on the last task to compare the methods under consideration. This metric is commonly known as the final average accuracy \cite{kirkpatrick_overcoming_2017,hsu_re-evaluating_2019}. For highlighting the benefits of our approach for retaining past knowledge, we also take into account the Backward Transfer (BT) metric~\cite{mai_online_2021,wang_comprehensive_2023}.

\section{Challenges of KD in OCL}
\label{sec:challenges}
In this section, we discuss unique challenges in OCL that make implementation of KD in this context laborious.

\subsection{Teacher Quality}
Given that incoming data can be seen by the model only once, it is uncertain whether the model has been fully trained at the end of each task. Consequently, taking a snapshot of the model at the end of the previous task may result in a suboptimal teacher. Such a teacher might hinder the student model's training for the subsequent task, leading to further degradation in the quality of teachers for the next task and an overall decline in performance. This problem is magnified when starting from a randomly initialized model, which is a common practice in OCL. Moreover, a model's performance on a specific task greatly depends on the difficulty of said task. Starting with a difficult task can lead to an especially low-quality teacher, further harming the distillation process.

Examples of such performance gaps are shown in Table~\ref{tab:teacher_quality1} with GSA~\cite{guo_dealing_2023}, a state-of-the-art approach. It can be observed that training offline leads to significantly higher performance than training online. Similarly, beginning training with an easy task induces superior performance on said task when compared with a hard task. Additional insights regarding the importance of teacher quality are given in Table~\ref{tab:teacher_quality2} where we show the impact of two distillation strategies on the final performances ER~\cite{rolnick_experience_2019}. Namely, we combine ER with a low-quality teacher that is a snapshot of the model at the end of the previous task. Similarly, we combine ER with a high-quality teacher that is a snapshot of a model trained for 5 epochs on the previous task. We use the training loss defined in Equation~\eqref{eq:loss_kd} after conducting a small hyper-parameter search on $\lambda$. It can be observed that while the impact of a low-quality teacher is limited, the impact of a higher-quality teacher is significant.
\begin{table}[t]
    \centering
    \caption{Accuracy of GSA~\cite{guo_dealing_2023} on the first task of CIFAR100 M=5k splited in 10 tasks, on different training scenarios. We train for 20 epochs for Offline CL, 1 epoch for Online CL.}
    \begin{tabular}{l|c}
        \toprule
        Training Scenario & Accuracy (\%) \\
        \midrule
        Offline CL       &     81.8 \\
        Online  CL       &     61.0 \\
        \midrule
        Online CL, Hard task    &   51.6  \\
        Online CL, Easy task    &   72.1   \\
        \bottomrule
    \end{tabular}
    \vspace{-1em}
    \label{tab:teacher_quality1}
\end{table}

\begin{table}[t]
    \centering
    \caption{Accuracy of ER~\cite{rolnick_experience_2019} using a low-quality teacher (snapshot of the model at the end of previous task), and a high-quality teacher (snapshot of a model trained for 5 epochs on previous task), on CIFAR100 M=5k splited in 2 tasks. We use $\lambda=0.01$ after conducting a small hyper-parameter search. Means and standard deviations over 5 runs are reported.}
    \begin{tabular}{l|c}
        \toprule
        Method & Accuracy (\%) \\
        \midrule
        ER                      &     49.0$\pm$4.6\\ 
        ER+low\ qual.\ teach.   &     50.7$\pm$4.3\\
        ER+high\ qual.\ teach.  &     54.6$\pm$3.3\\
        \bottomrule
    \end{tabular}
    \vspace{-1em}
    \label{tab:teacher_quality2}
\end{table}

\subsection{Teacher Quantity}
One strategy for applying KD to CL requires taking a snapshot of the model at the end of each task~\cite{rannen2017encoder, ahn_ss-il_2021, hou2018lifelong}. Each snapshot then serves as a teacher for the respective task and is incorporated into the distillation loss. Naturally, this requires storing a copy of the model per task which can be problematic for a large number of tasks, even in standard CL. We emphasize that memory consumption is crucial to OCL because it is presumed that only a small fraction of data can be retained, and all other incoming data is discarded post-usage. Dealing with a growing quantity of teachers is unrealistic and contradicts the implicit storage constraint of the online setup.

To circumvent the issue of continuously increasing teacher numbers, one might consider using just the snapshot from the most recent task as a teacher. However, this solution is also unsatisfactory as this teacher should encapsulate the knowledge from all previous tasks, which is especially complex for long task sequences.

\subsection{Unknown Task Boundaries}
\begin{figure}[t!]
\begin{center}
\centerline{\includegraphics[width=0.7\columnwidth]{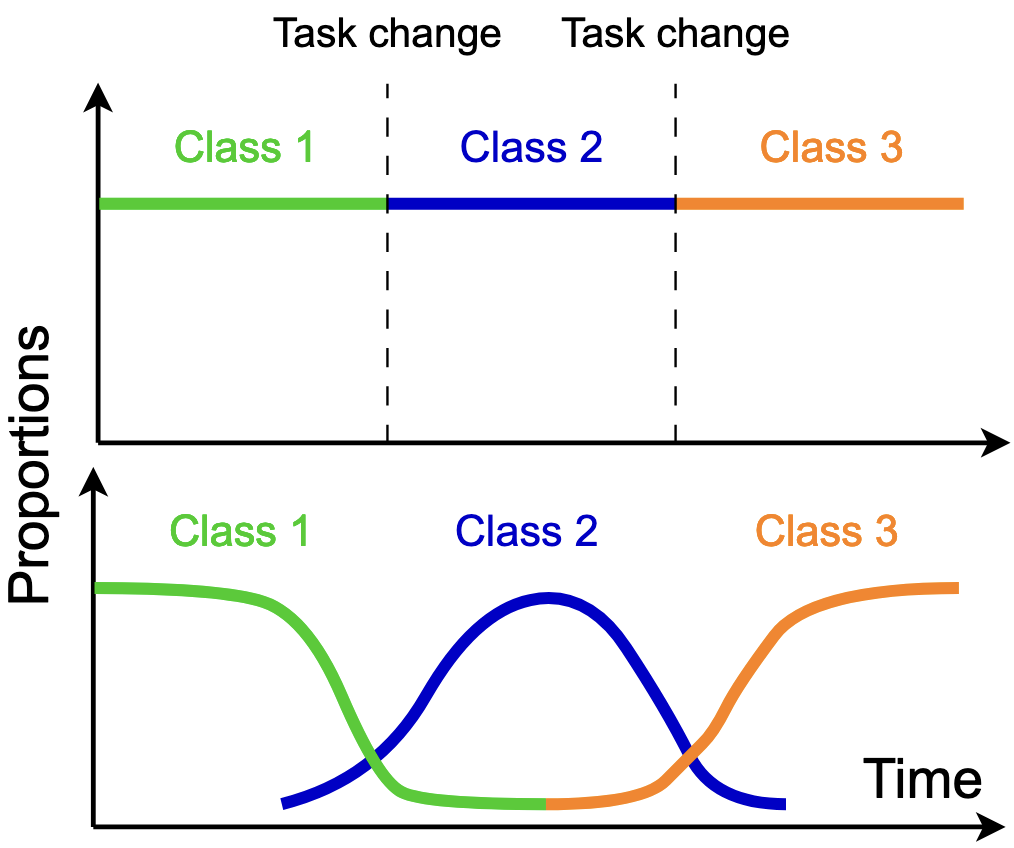}}
\caption{Illustration of the \textit{blurry} boundary setting (bottom row) in opposition to the \textit{clear} boundary setting (top row). Detecting task change in the case of \textit{blurry} is not trivial.}
\vspace{-1em}
\label{fig:knowing_task_boundaries}
\end{center}
\end{figure}

Most distillation strategies in CL rely on task boundaries information to select the best teachers for distillation. In offline CL, this information is easily available. However in OCL, pinpointing the exact moment of task change is not guaranteed. Figure~\ref{fig:knowing_task_boundaries} illustrates a more realistic scenario where transitions occur progressively, making the determination of the ideal snapshot moment challenging. Choosing a suboptimal teacher can also compromise the quality of distillation.

\section{Methodology}
\subsection{Motivations}

As mentioned in previous sections, KD has been underutilized in OCL. The main reason is that most KD strategies draw inspiration from offline CL where the teacher is typically frozen at the conclusion of the previous task. However, relying on a frozen teacher in OCL can be problematic due to unknown task boundaries and concerns regarding teacher quality. Moreover, a static teacher from the previous task will set an upper limit on the student's learning potential. Consequently, the student is unable to enhance performance on the previous task while mastering the current one. In other words, a simple teacher discourages backward transfer.

To tackle this limitation, we propose the use of an evolving teacher. Contrary to a fixed teacher, the weights of an evolving teacher are updated throughout the training process. This approach allows the teacher to continually improve and not hinder the student's progression. A student learning from an evolving teacher can consistently refine their performance on preceding tasks, thereby promoting backward transfer. Additionally, this kind of teacher eliminates the need for the knowledge of task boundaries. In this paper, we take advantage of an Exponential Moving Average (EMA) of the current model as the evolving teacher and design a novel MKD teacher-dependent weighting scheme for adapting MKD to OCL. While EMA can efficiently solve previously described challenges, its applications to OCL is still in its infancy.


\subsection{Momentum Knowledge Distillation}

We propose a new scheme to leverage Momentum Knowledge Distillation~\cite{he_momentum_2020} (MKD) with an evolving teacher. In this distillation strategy, the teacher architecture mirrors that of the student and its weights are computed as an Exponential Moving Average of the student parameters. The EMA weights are computed online according to the update parameters $\alpha$ such that:
\begin{equation}
    \theta_{\alpha}(t)=\alpha*\theta(t) + (1-\alpha)*\theta_{\alpha}(t-1),
\end{equation}
where $\theta(t)$ represents the student's model parameters at time $t$. The teacher, parameterized by $\theta_{\alpha}$, is represented as $\mathcal{T}_{\alpha}$.

\subsection{Rethinking MKD}
\begin{figure}[t!]
\begin{center}
\centerline{\includegraphics[width=0.6\columnwidth]{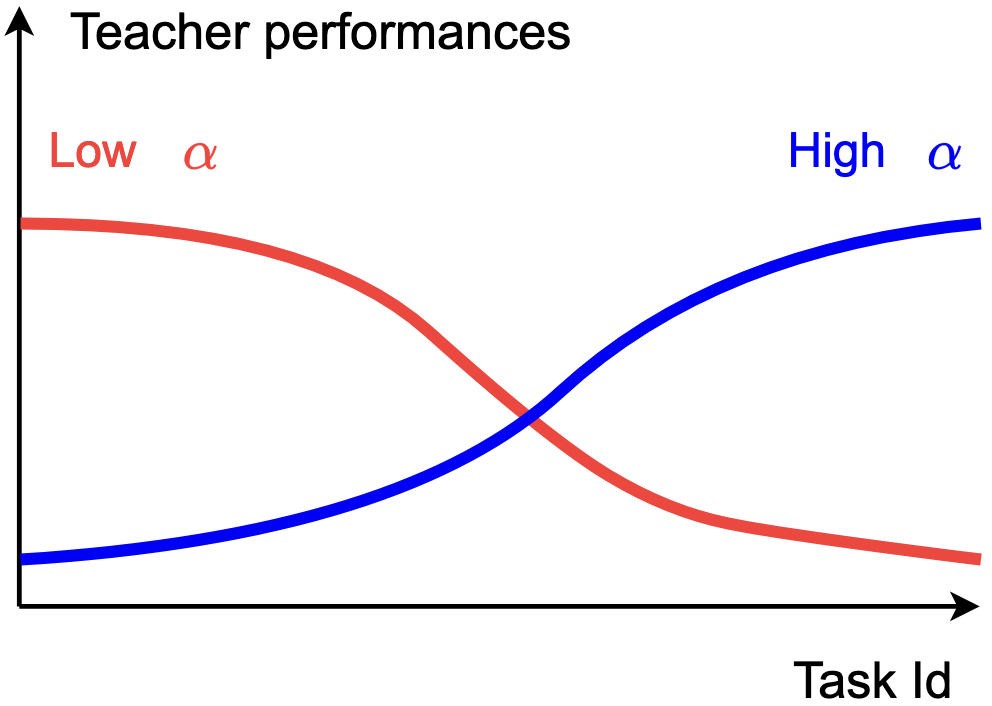}}
\caption{Impact of $\alpha$ on the plasticity-stability trade-off. Lower $\alpha$ values imply a stable teacher with high performances on old tasks. Higher $\alpha$ implies a plastic teacher, with high performances on new tasks.}
\vspace{-1em}
\label{fig:plasticity_stability_alpha}
\vspace{-0.2cm}
\end{center}
\end{figure}

\paragraph{Plasticity-stability Control} When designing CL methods, it is common to address the plasticity-stability trade-off~\cite{wang_comprehensive_2023}. Usually, the application of distillation augments the model's stability at the expense of its plasticity. Using Momentum Knowledge Distillation gives a precise control over this trade-off through the parameter $\alpha$. A lower value of $\alpha$ would make the teacher update slower and remember longer timelines, making it retain longer timelines but offering scant knowledge on the current task. A high value of $\alpha$ would help the student learn the current task but with limited insight of previous tasks. In other words, a higher value of $\alpha$ emphasizes plasticity over stability whereas a lower value of $\alpha$ encourages stability over plasticity. This plasticity-stability control characteristic is illustrated in Figure~\ref{fig:plasticity_stability_alpha}. We make concrete usage of this property by designing a teacher-dependent weighting scheme in our model learning.

\paragraph{Model Learning}
We formulate our loss term using an EMA teacher as described in equation \ref{eq:loss_kd}.
\begin{equation}
\begin{aligned}
\label{eq:loss_kd}
\mathcal{L}(X,Y)=&\mathcal{L}_{CE}(X,Y) + \\ 
                    &\lambda_\alpha * KL(\mathcal{T}_{\alpha}(X)/\tau, S(X)/\tau),
\end{aligned}
\end{equation}
where $\mathcal{L}_{CE}$ the Cross-Entropy function, $\lambda_\alpha$ a weighting hyper-parameter depending on $\alpha$, $S$ the student model, $(X,Y)$ the data-label pairs, $KL$ the Kullback–Leibler divergence and $\tau$ the distillation temperature. We further introduce multiview distillation, by making use of a data augmentation procedure $Aug(.)$ and propose to minimize $\mathcal{L}_{MKD}$ defined in Equation~\ref{eq:loss_mkd}.
\begin{equation}
\label{eq:loss_mkd}
\begin{aligned}
\mathcal{L}_{MKD}(X,Y)=& \mathcal{L}_{CE}(\hat{X},Y) \\ + &\frac{\lambda_\alpha}{2}KL(\mathcal{T}_{\alpha}(X), S(\hat{X}))\\ + & \frac{\lambda_\alpha}{2}KL(\mathcal{T}_{\alpha}(\hat{X}), S(\hat{X})),
\end{aligned}
\end{equation}
where $\hat{X} = Aug(X)$.

The only hyper-parameter is $\alpha$. In Section~\ref{sec:discussion}, we give details on how to efficiently choose $\alpha$ and how to express the teacher-dependent weighting parameter $\lambda_\alpha$. 
Additionally, the simplicity of this process allows for seamless adaptation to existing methods. We provide a PyTorch-like~\cite{paszke2019pytorch} pseudo-code that outlines the strategy for integrating our proposed MKD into other training procedures, as can be found in Algorithm~\ref{code:pseudo_code}.

\begin{algorithm}[t]
\tiny
\begin{minted}{python}
for x, y in dataloader:
  # Baseline loss
  loss_baseline = criterion_baseline(model, x, y)
  loss = loss_baseline
  
  # Proposed loss
  x_aug = transform(x)  # data augmentation
  l_stu1 = model(x) # logits student
  l_stu2 = model(x_aug) # logits studenwt
  l_tea = teacher(x_aug) # logits teacher
  loss_ce = cross_entropy(x_aug, y)
  loss_d1 = kl_div(softmax(l_stu1/t), softmax(l_tea/t)) # temperature t 
  loss_d2 = kl_div(softmax(l_stu2/t), softmax(l_tea/t)) 
  loss_dist = (loss_d1 + loss_d2)/2  # Eq. 3
  loss += loss_ce + lam*loss_dist
  
  optim.zero_grad()
  loss.backward()
  optim.step()
  update_ema()
\end{minted}
 \caption{PyTorch-like pseudo-code of our loss to integrate to other baselines.}
\label{code:pseudo_code}
\end{algorithm}

In this pseudo-code, we have omitted a memory buffer for simplicity. Nonetheless, the training procedure remains consistent, using a batch combining stream and memory data.

\paragraph{Model Estimation}
As introduced in the plasticity-stability control section, the knowledge of the teacher and student pertains to different tasks. The student is inclined towards the current task whereas the teacher excels in past tasks. Solely relying on the teacher's or student's weights for inference may not yield optimal performances. Consequently, we introduce a new model estimation strategy that necessitates minimal extra computation. We compute the final model parameters $\theta^\star$ as the average of teacher and student weights such that $\theta^\star=\frac{\theta_S + \theta_T}{2}$, where $\theta_S$ and $\theta_T$ denote the parameters of the student and teacher, respectively. A similar model estimation strategy has been employed in conventional image classification~\cite{tarvainen2017mean}. We show in Section~\ref{sec:ablation_studies} that this strategy can enhance performance.

\section{Experiments}
\subsection{Implementation Details}
For each method, we use random retrieval and reservoir sampling~\cite{vitter_random_1985} for memory management. We use a full ResNet18~\cite{he_deep_2015} (untrained) for every method. For all baselines, we perform a small hyperparameter search on CIFAR100, M=5k, applying the determined parameters across other configurations. More details are given in the Appendix. We use the same hyperparameters when incorporating our loss. Throughout the training process, the streaming batch size is set to 10, and data retrieval from memory is capped at 64. Data augmentation includes random flip, grayscale, color jitter, and random crop. The \textit{blurry} datasets are created following the code given in~\cite{michel_learning_2023} with a scale of 500. Some methods require task boundary inference to be adapted to the \textit{blurry} setting, which is detailed in Appendix. The temperature $\tau$ designated for KD is $4$. For MKD, we use $\alpha=0.01$ and $\lambda_\alpha=5.5$ accordingly for every method. For more details regarding experiments, please refer to the Appendix.

\subsection{Baselines}
To show the efficiency of our proposed approach, we integrate our approach as described in our pseudo code into several baselines and the state-of-the-art methods in OCL.\\
\textbf{ER}~\cite{rolnick_experience_2019}: A basic memory based method leveraging a Cross-Entropy loss and a replay buffer. \textbf{DER++}~\cite{buzzega_dark_2020}: A replay-based approach doing distillation of old stored logits with using task boundaries. \textbf{ER-ACE}~\cite{caccia_new_2022}: A replay-based method using an Asymmetric Cross Entropy to overcome feature drift. \textbf{DVC}~\cite{gu_not_2022}: A replay-based approach leveraging consistency between image views in addition to minimizing cross entropy. \textbf{OCM}~\cite{guo_online_2022}: A replay-based method maximizing mutual information between old and new samples representation. \textbf{GSA}~\cite{guo_dealing_2023}: A replay-based method dealing with cross-task class discrimination with a redefined loss objective using Gradient Self Adaptation. \textbf{PCR}~\cite{lin2023pcr}: A replay-based method leveraging a proxy-based contrastive loss for OCL. \textbf{Temp. Ens.}~\cite{soutif--cormerais_improving_2023} leverages temporal ensembles in OCL. Specifically, the authors use the EMA of the current model for inference, although it is not used for distillation. We report the performances of Temp. Ens. combined with ER for comparison. \textbf{SDP}~\cite{koh_online_2023} uses a hypo-exponential evolving teacher. We report the performances of SDP combined with ER for comparison.

For reproducibility, we re-implemented the methods mentioned above and make the code public.

\subsection{Experimental Results}

\paragraph{Clear Boundary Setting}
To demonstrate the effectiveness of our approach, we applied the procedure described to all the considered baselines and compared the performances. Average accuracy at the end of training for the \textit{clear} setting is displayed in Table~\ref{tab:clear_setting_all}. It can be observed that for most of the considered methods, datasets and memory sizes, applying our procedure improves performance. In most cases, this gain in performance is significant. Specifically, the combinations \textit{GSA + ours} and \textit{OCM + ours} have the potential to surpass the current state-of-the-art methods. Additionally, the standard deviation is also significantly reduced when applying our approach, showing that the use of a momentum teacher can help stabilizing the training procedure. More interestingly, the introduction of our distillation procedure can enhance performance, even if distillation is already incorporated in the method (e.g., \textit{DER++}).

\begin{table*}[t]
    \centering
    \caption{Final average accuracy (\%) for the \textit{clear} boundary setting at the end of training for considered baselines, with and without our additional MKD procedure. Results are displayed for different datasets and memory sizes. Displayed values are the mean and standard deviation computed over 5 runs.}
    \resizebox{\textwidth}{!}{
    \begin{tabular}{l|lll|lll|lll|lll} \multicolumn{1}{c}{Dataset} & \multicolumn{3}{c}{CIFAR10} & \multicolumn{3}{c}{CIFAR100} & \multicolumn{3}{c}{Tiny-IN} & \multicolumn{3}{c}{ImageNet100} \\
    \hline
\multicolumn{1}{c}{Memory Size M} & \multicolumn{1}{c}{200} & \multicolumn{1}{c}{500} & \multicolumn{1}{c}{1000} & \multicolumn{1}{c}{1000} & \multicolumn{1}{c}{2000} & \multicolumn{1}{c}{5000} & \multicolumn{1}{c}{2000} & \multicolumn{1}{c}{5000} & \multicolumn{1}{c}{10000} & \multicolumn{1}{c}{2000} & \multicolumn{1}{c}{5000} & \multicolumn{1}{c}{10000}\\
\hline\hline
ER~\text{[NeurIPS`19]}       &  46.33{\scriptsize±2.42} &  55.73{\scriptsize±2.04} &    62.99{\scriptsize±2.1} &  23.0{\scriptsize±0.8}   &  31.55{\scriptsize±1.27} &  38.05{\scriptsize±1.08} &  11.39{\scriptsize±0.75} &  18.97{\scriptsize±1.16} &  21.52{\scriptsize±3.37} &    19.06{\scriptsize±0.9} &   29.74{\scriptsize±1.34} &   36.72{\scriptsize±1.09} \\
ER + SDP                     &  47.78{\scriptsize±2.29} &  58.85{\scriptsize±1.38} &   65.93{\scriptsize±2.15} &  28.48{\scriptsize±1.18} &  35.21{\scriptsize±0.73} &  41.19{\scriptsize±1.11} &  15.32{\scriptsize±0.47} &  23.22{\scriptsize±0.31} &   26.97{\scriptsize±1.1} &   22.95{\scriptsize±0.13} &   32.12{\scriptsize±0.23} &   35.64{\scriptsize±0.25} \\
ER + Temp. Ens. & 48.80{\scriptsize±2.60} & 62.10{\scriptsize±1.70} & 70.20{\scriptsize±0.20} & 30.00{\scriptsize±0.70} & 37.90{\scriptsize±0.90} & 46.80{\scriptsize±0.90} & 16.30{\scriptsize±0.50} & 23.60{\scriptsize±0.40}& 30.00{\scriptsize±0.50} &  20.40{\scriptsize±0.50} &  33.12{\scriptsize±1.70} &  40.10{\scriptsize±1.15} \\
ER + ours                    &  \textbf{57.54{\scriptsize±2.55}} &  \textbf{68.48{\scriptsize±0.92}} &   \textbf{74.33{\scriptsize±0.68}} &  \textbf{38.5{\scriptsize±0.5}}   &  \textbf{45.2{\scriptsize±0.2}}   &  \textbf{52.10{\scriptsize±0.5}} &  \textbf{23.95{\scriptsize±0.65}} &  \textbf{32.22{\scriptsize±0.88}} &  \textbf{38.27{\scriptsize±0.18}} &   \textbf{30.67{\scriptsize±0.46}} &    \textbf{39.7{\scriptsize±1.07}} &   \textbf{44.92{\scriptsize±0.98}} \\
\hline
DER++~\text{[NeurIPS`20]}    &  47.07{\scriptsize±0.97} &  55.53{\scriptsize±1.05} &   58.51{\scriptsize±0.68} &   22.8{\scriptsize±1.8} &  25.89{\scriptsize±1.46} &   25.71{\scriptsize±2.4} &   3.89{\scriptsize±0.64} &   4.28{\scriptsize±0.51} &   4.16{\scriptsize±0.32} &   15.36{\scriptsize±3.04} &   19.19{\scriptsize±1.55} &   20.48{\scriptsize±4.67} \\
DER++ + ours                 &  \textbf{53.63{\scriptsize±2.18}} &  \textbf{63.95{\scriptsize±0.86}} &   \textbf{68.84{\scriptsize±1.15}} &  \textbf{32.1{\scriptsize±0.5}} &  \textbf{37.97{\scriptsize±0.92}} & \textbf{41.97{\scriptsize±1.53}} &  \textbf{17.08{\scriptsize±1.43}} &  \textbf{15.64{\scriptsize±4.64}} & \textbf{ 13.69{\scriptsize±3.36}} &   \textbf{26.18{\scriptsize±1.07}} &   \textbf{33.85{\scriptsize±0.98}} &   \textbf{38.22{\scriptsize±1.84}} \\
\hline
ERACE~\text{[ICLR`22]}       &  44.77{\scriptsize±3.18} &  52.65{\scriptsize±1.37} &   61.45{\scriptsize±1.47} &  27.4{\scriptsize±0.6} &  32.88{\scriptsize±0.63} &  39.61{\scriptsize±0.53} &  14.79{\scriptsize±0.95} &  22.25{\scriptsize±1.69} &  26.64{\scriptsize±0.91} &   27.16{\scriptsize±0.57} &   32.88{\scriptsize±0.83} &   39.14{\scriptsize±0.35} \\
ERACE + ours                 &  \textbf{58.99{\scriptsize±1.36}} &  \textbf{65.94{\scriptsize±0.49}} &   \textbf{69.78{\scriptsize±0.96}} &  \textbf{37.0{\scriptsize±0.7}} &  \textbf{42.92{\scriptsize±0.79}} &  \textbf{48.73{\scriptsize±1.29}} &  \textbf{22.21{\scriptsize±0.87}} &  \textbf{31.13{\scriptsize±0.41}} &  \textbf{35.54{\scriptsize±0.43}} &   \textbf{33.59{\scriptsize±0.99}} &   \textbf{41.93{\scriptsize±0.64}} &   \textbf{47.16{\scriptsize±0.89}} \\
\hline
DVC~\text{[CVPR`22]}         &  48.08{\scriptsize±4.27} &  58.72{\scriptsize±2.03} &   61.11{\scriptsize±2.97} &  18.66{\scriptsize±2.54} &   22.73{\scriptsize±2.9} &  28.47{\scriptsize±3.95} &    2.04{\scriptsize±0.8} &   1.47{\scriptsize±0.49} &   1.54{\scriptsize±0.79} &   14.54{\scriptsize±5.15} &   21.88{\scriptsize±3.45} &    28.5{\scriptsize±2.93} \\
DVC + ours                   &  \textbf{50.53{\scriptsize±4.35}} &  \textbf{62.62{\scriptsize±1.84}} &   \textbf{69.52{\scriptsize±0.84}} &  \textbf{27.42{\scriptsize±3.14}} &  \textbf{35.95{\scriptsize±1.71}} &  \textbf{42.45{\scriptsize±2.45}} &   \textbf{9.41{\scriptsize±1.43}} &  \textbf{12.03{\scriptsize±3.83}} &  \textbf{13.44{\scriptsize±3.84}} &   \textbf{18.75{\scriptsize±1.96}} &    \textbf{29.64{\scriptsize±3.6}} &    \textbf{38.0{\scriptsize±2.94}} \\
\hline
OCM~\text{[ICML`22]}         &  59.58{\scriptsize±1.43} &  68.46{\scriptsize±0.79} &   72.79{\scriptsize±2.36} &   29.3{\scriptsize±1.55} &   36.7{\scriptsize±0.58} &  41.87{\scriptsize±1.52} &  19.58{\scriptsize±0.63} &  27.85{\scriptsize±1.03} &  32.56{\scriptsize±1.37} &    28.7{\scriptsize±0.92} &   37.37{\scriptsize±1.11} &   41.86{\scriptsize±1.14} \\
OCM + ours                   &  \textbf{67.02{\scriptsize±3.14}} &  \textbf{75.14{\scriptsize±0.75}} &   \textbf{79.33{\scriptsize±0.55}} &  \textbf{38.21{\scriptsize±0.62}} &  \textbf{45.51{\scriptsize±0.94}} &  \textbf{51.24{\scriptsize±0.81}} &  \textbf{23.07{\scriptsize±0.37}} &  \textbf{31.82{\scriptsize±0.72}} &  \textbf{37.46{\scriptsize±0.95}} &   \textbf{28.87{\scriptsize±1.85}} &   \textbf{38.26{\scriptsize±1.06}} &   \textbf{44.24{\scriptsize±0.55}} \\
\hline
GSA~\text{[CVPR`23]}         &  48.9{\scriptsize±3.38}  &  61.45{\scriptsize±1.95} &   67.63{\scriptsize±1.24} &  29.68{\scriptsize±1.54} &  36.96{\scriptsize±0.79} &  45.86{\scriptsize±1.89} &  15.77{\scriptsize±0.72} &   22.48{\scriptsize±0.4} &  28.46{\scriptsize±1.85} &   24.29{\scriptsize±0.59} &   33.47{\scriptsize±1.18} &   40.18{\scriptsize±0.93} \\
GSA + SDP                    &  47.39{\scriptsize±1.76} &  60.61{\scriptsize±3.43} &   67.17{\scriptsize±1.41} &  26.56{\scriptsize±3.03} &  34.78{\scriptsize±3.72} &  44.53{\scriptsize±1.07} &  11.71{\scriptsize±2.69} &    16.1{\scriptsize±6.3} &  25.92{\scriptsize±3.05} &   27.7{\scriptsize±10.69} &   43.85{\scriptsize±1.42} &   51.39{\scriptsize±1.07}\\
GSA + ours                   &  \textbf{57.66{\scriptsize±4.11}} &  \textbf{68.16{\scriptsize±0.85}} &   \textbf{75.08{\scriptsize±1.14}} &  \textbf{40.1{\scriptsize±1.0}} &  \textbf{48.23{\scriptsize±0.78}} &   \textbf{56.15{\scriptsize±0.6}} &  \textbf{23.14{\scriptsize±0.44}} &  \textbf{32.38{\scriptsize±1.28}} &  \textbf{38.78{\scriptsize±0.65}} &    \textbf{33.4{\scriptsize±0.85}} &   \textbf{44.99{\scriptsize±0.46}} &   \textbf{52.41{\scriptsize±0.59}} \\
\hline
PCR~\text{[CVPR`23]}         &  52.2{\scriptsize±0.66}  &  60.61{\scriptsize±2.23} &  61.66{\scriptsize±13.86} &  30.68{\scriptsize±0.81} &  38.63{\scriptsize±1.01} &  45.27{\scriptsize±0.78} &  12.47{\scriptsize±3.56} &  20.41{\scriptsize±2.84} &  23.85{\scriptsize±4.21} &   19.89{\scriptsize±6.24} &   31.35{\scriptsize±3.01} &    36.99{\scriptsize±4.7} \\
PCR+ours                     &  \textbf{55.83{\scriptsize±2.35}} &  \textbf{67.03{\scriptsize±1.33}} &   \textbf{73.47{\scriptsize±0.53}} &  \textbf{35.27{\scriptsize±0.47}} &  \textbf{44.95{\scriptsize±0.44}} &  \textbf{54.44{\scriptsize±0.46}} &  \textbf{17.14{\scriptsize±0.48}} &  \textbf{29.05{\scriptsize±0.55}} &  \textbf{36.65{\scriptsize±0.90}} &   \textbf{25.42{\scriptsize±0.54}} &   \textbf{39.50{\scriptsize±1.00}} &   \textbf{49.66{\scriptsize±1.78}} \\
\hline
    \end{tabular}}
    
    \label{tab:clear_setting_all}
\end{table*}
\begin{table*}[t]
    \centering
    \caption{Final average accuracy (\%) for the \textit{blurry} boundary setting at the end of training for considered baselines, with and without our additional MKD procedure. Results are displayed for different datasets and memory sizes. Displayed values are the mean and standard deviation computed over 5 runs.}
    \resizebox{\textwidth}{!}{
    \begin{tabular}{l|lll|lll|lll|lll} \multicolumn{1}{c}{Dataset} & \multicolumn{3}{c}{CIFAR10} & \multicolumn{3}{c}{CIFAR100} & \multicolumn{3}{c}{Tiny-IN} & \multicolumn{3}{c}{ImageNet100} \\
    \hline
\multicolumn{1}{c}{Memory Size M} & \multicolumn{1}{c}{200} & \multicolumn{1}{c}{500} & \multicolumn{1}{c}{1000} & \multicolumn{1}{c}{1000} & \multicolumn{1}{c}{2000} & \multicolumn{1}{c}{5000} & \multicolumn{1}{c}{2000} & \multicolumn{1}{c}{5000} & \multicolumn{1}{c}{10000} & \multicolumn{1}{c}{2000} & \multicolumn{1}{c}{5000} & \multicolumn{1}{c}{10000}\\
\hline\hline
ER~\text{[NeurIPS`19]}       &  44.78{\scriptsize±6.22} &   54.1{\scriptsize±4.54} &  64.17{\scriptsize±1.89} &  24.76{\scriptsize±1.33} &  31.56{\scriptsize±1.73} &    39.1{\scriptsize±1.0} &   11.88{\scriptsize±1.5} &  19.76{\scriptsize±1.67} &  25.71{\scriptsize±1.29} &   15.18{\scriptsize±1.51} &   24.88{\scriptsize±1.27} &   31.44{\scriptsize±2.18} \\
ER + SDP                     &  47.64{\scriptsize±1.66} &  60.04{\scriptsize±1.53} &  65.89{\scriptsize±0.84} &  28.66{\scriptsize±1.61} &   36.07{\scriptsize±1.5} &  41.65{\scriptsize±1.47} &   15.98{\scriptsize±1.6} &  23.91{\scriptsize±1.48} &  28.92{\scriptsize±0.84} &   14.26{\scriptsize±1.47} &    5.14{\scriptsize±3.14} &    5.15{\scriptsize±2.16} \\
ER + Temp. Ens. & 51.58{\scriptsize±2.17} & 64.12{\scriptsize±0.57} &  70.90{\scriptsize±0.90} & 31.78{\scriptsize±0.64} & 39.32{\scriptsize±0.86} & 46.84{\scriptsize±0.65} & 16.63{\scriptsize±0.80} & 25.37{\scriptsize±1.11} &  32.07{\scriptsize±0.50} & 14.45{\scriptsize±0.33} & 24.57{\scriptsize±0.81} &  27.09{\scriptsize±0.68} \\
ER + ours                    &  \textbf{56.69{\scriptsize±1.9}}  &  \textbf{69.15{\scriptsize±1.37}} &  \textbf{74.06{\scriptsize±1.01}} &  \textbf{38.38{\scriptsize±0.86}} &  \textbf{45.47{\scriptsize±0.63}} &  \textbf{51.79{\scriptsize±0.22}} &  \textbf{25.08{\scriptsize±0.64}} &  \textbf{33.22{\scriptsize±0.64}} &   \textbf{38.63{\scriptsize±0.8}} &   \textbf{26.03{\scriptsize±0.76}} &   \textbf{36.67{\scriptsize±0.65}} &   \textbf{42.64{\scriptsize±0.87}} \\
\hline
DER++~\text{[NeurIPS`20]}    &  47.28{\scriptsize±2.03} &  55.83{\scriptsize±2.45} &  59.37{\scriptsize±1.93} &   23.4{\scriptsize±1.54} &   27.91{\scriptsize±1.3} &  29.31{\scriptsize±1.83} &   15.99{\scriptsize±0.9} &  20.34{\scriptsize±1.22} &  21.36{\scriptsize±0.81} &    3.65{\scriptsize±1.38} &    3.98{\scriptsize±1.53} &    4.22{\scriptsize±1.66} \\
DER++ + ours                 &  \textbf{54.21{\scriptsize±3.11}} &  \textbf{63.83{\scriptsize±1.83}} &   \textbf{69.06{\scriptsize±1.7}} &  \textbf{31.17{\scriptsize±0.81}} &  \textbf{38.44{\scriptsize±1.15}} &  \textbf{42.72{\scriptsize±0.81}} &  \textbf{21.93{\scriptsize±0.74}} &   \textbf{28.7{\scriptsize±0.55}} &  \textbf{32.58{\scriptsize±1.51}} &   \textbf{20.69{\scriptsize±1.09}} &   \textbf{27.37{\scriptsize±0.93}} &   \textbf{30.25{\scriptsize±1.01}} \\
\hline
ERACE~\text{[ICLR`22]}       &  50.44{\scriptsize±1.37} &   56.5{\scriptsize±1.77} &   62.92{\scriptsize±1.5} &  27.69{\scriptsize±1.45} &  32.98{\scriptsize±0.81} &  40.12{\scriptsize±1.05} &  19.04{\scriptsize±0.88} &  25.27{\scriptsize±1.27} &  30.05{\scriptsize±1.67} &   15.84{\scriptsize±0.82} &    24.49{\scriptsize±0.3} &   31.74{\scriptsize±0.97} \\
ERACE + ours                 &  \textbf{59.36{\scriptsize±3.15}} &  \textbf{66.25{\scriptsize±1.82}} &   \textbf{70.74{\scriptsize±0.8}} &  \textbf{38.04{\scriptsize±0.93}} &  \textbf{43.75{\scriptsize±0.46}} &  \textbf{50.35{\scriptsize±0.48}} &   \textbf{25.85{\scriptsize±0.7}} &  \textbf{33.14{\scriptsize±0.79}} &  \textbf{37.68{\scriptsize±0.57}} &   \textbf{26.14{\scriptsize±0.64}} &   \textbf{35.61{\scriptsize±1.06}} &   \textbf{42.67{\scriptsize±0.81}} \\
\hline
DVC~\text{[CVPR`22]}         &  46.05{\scriptsize±5.23} &   58.73{\scriptsize±2.4} &  58.78{\scriptsize±5.83} &  22.46{\scriptsize±1.91} &  26.98{\scriptsize±3.13} &  29.46{\scriptsize±2.39} &  10.64{\scriptsize±1.31} &   15.48{\scriptsize±2.1} &  15.81{\scriptsize±1.76} &    2.97{\scriptsize±0.65} &    5.34{\scriptsize±2.49} &    8.38{\scriptsize±4.58} \\
DVC + ours                   &  \textbf{49.04{\scriptsize±2.95}} &  \textbf{61.95{\scriptsize±1.81}} &  \textbf{69.25{\scriptsize±0.78}} &   \textbf{27.1{\scriptsize±2.11}} & \textbf{35.76{\scriptsize±2.27}} &  \textbf{41.99{\scriptsize±3.31}} &  \textbf{12.45{\scriptsize±2.19}} &  \textbf{22.15{\scriptsize±1.42}} &  \textbf{24.14{\scriptsize±3.63}} &    \textbf{6.53{\scriptsize±1.12}} &   \textbf{15.32{\scriptsize±1.28}} &   \textbf{5.42{\scriptsize±1.35}} \\
\hline
OCM~\text{[ICML`22]}         &  43.66{\scriptsize±2.59} &  47.63{\scriptsize±2.68} &  51.08{\scriptsize±2.66} &  25.16{\scriptsize±0.76} &  32.96{\scriptsize±1.21} &  38.14{\scriptsize±1.11} &  18.57{\scriptsize±0.37} &  26.82{\scriptsize±0.86} &  31.21{\scriptsize±0.55} &   \textbf{26.61{\scriptsize±1.02}} &   \textbf{36.36{\scriptsize±0.48}} &    \textbf{41.92{\scriptsize±0.9}} \\
OCM + ours                   &  \textbf{67.66{\scriptsize±0.49}} &   \textbf{74.9{\scriptsize±0.98}} &  \textbf{78.61{\scriptsize±0.43}} &  \textbf{36.64{\scriptsize±0.47}} &  \textbf{44.63{\scriptsize±1.12}} &  \textbf{51.41{\scriptsize±0.71}} &   \textbf{24.77{\scriptsize±0.2}} &   \textbf{33.01{\scriptsize±1.1}} &  \textbf{39.39{\scriptsize±0.89}} &   25.52{\scriptsize±1.27} &   34.42{\scriptsize±0.97} &    39.07{\scriptsize±0.8} \\
\hline
PCR~\text{[CVPR`23]}         &  53.43{\scriptsize±2.2} &  60.67{\scriptsize±3.29} &  69.13{\scriptsize±0.66} &   30.9{\scriptsize±2.06} &  38.63{\scriptsize±0.26} &  45.97{\scriptsize±1.18} &   16.0{\scriptsize±2.35} &  22.02{\scriptsize±3.21} &   28.9{\scriptsize±3.73} &    9.77{\scriptsize±4.75} &   16.55{\scriptsize±7.91} &   27.86{\scriptsize±5.46} \\
PCR+ours                     &  \textbf{57.55{\scriptsize±1.4}} &   \textbf{67.03{\scriptsize±2.0}} &   \textbf{74.0{\scriptsize±0.91}} &   \textbf{35.6{\scriptsize±0.66}} &  \textbf{44.95{\scriptsize±0.42}} &  \textbf{54.87{\scriptsize±0.39}} &  \textbf{17.33{\scriptsize±1.28}} &   \textbf{29.58{\scriptsize±0.6}} &  \textbf{38.02{\scriptsize±1.64}} &   \textbf{22.51{\scriptsize±0.96}} &   \textbf{34.53{\scriptsize±0.57}} &   \textbf{44.28{\scriptsize±0.68}} \\
\hline
    \end{tabular}}
    \label{tab:blurry_all}
\end{table*}

\paragraph{Blurry Boundary Setting}
To further demonstrate the capabilities of MKD, we also conducted experiments with \textit{blurry} task boundaries. Average accuracy at the end of training is shown in Table \ref{tab:clear_setting_all}. However, we did not implement GSA in this context since it requires knowledge of the exact class-task relationships and is not easily adaptable to this setup. Additionally, we inferred task boundaries for OCM as it is required to apply the method. Details on how the task boundaries are inferred in this setup are given in Appendix. Similar to the \textit{clear} boundary setting, incorporating MKD as per our procedure can significantly enhance performance. This performance gain becomes even more pronounced when the original method experiences a drop in effectiveness due to the challenging nature of the setting. For example, OCM performances on CIFAR100 M=5k drop from $41.87\%$ to $38.14\%$ while \textit{OCM + ours} performances remain stable around $51.4\%$.

\paragraph{Comparison with SDP}
SDP~\cite{koh_online_2023} uses a hypo-exponential evolving teacher, akin to our approach. While initially proposed as a standalone method, SDP can be combined with existing techniques. We integrated SDP with \textit{ER} and \textit{GSA}, and results in Table~\ref{tab:clear_setting_all} reveal that, although SDP enhances \textit{ER}, \textit{ER + SDP} performs less effectively than \textit{ER + ours}. Additionally, for \textit{GSA}, the inclusion of SDP leads to decreased performance, confirming MKD's superiority over SDP. Computationally, as SDP operates in representation space, it demands more resources compared to MKD, which is computed in logit space. Further details on the computational constraints are provided in the Appendix. The introduction of SDP has a more substantial impact on the time consumption of \textit{ER} and \textit{GSA} than MKD.

\subsection{Ablation Studies}
\label{sec:ablation_studies}

\paragraph{Impact of the Final Weight Estimation}
To demonstrate the impact of averaging weights from the teacher and the student, we experimented using either the teacher or the student exclusively for inference. Results are displayed for \textit{ER} on Table \ref{tab:ablation}. In both cases, employing solely the student or the teacher results in inferior performance compared to using their averaged weights, with a minimum drop in accuracy of $0.5\%$. Additionally, the teacher performs worse than the student, which can be due to the fact that for remembering enough from past tasks, the teacher update must be quite slow. In that sense, the teacher might perform worse overall but improve the students' stability.

\begin{table}[t]
    \centering
    \caption{Final average accuracy (\%) on CIFAR100, \textit{clear} boundary setting, for \textit{ER + ours} and varying memory sizes. \textit{Student} corresponds to the student performance and \textit{teacher} to the teacher performance. \textit{no aug} corresponds using the distillation loss with a single view as defined in Section~\ref{sec:ablation_studies}. Mean and standard deviations over 5 runs are displayed.}
    \resizebox{0.9\columnwidth}{!}{
    \begin{tabular}{l|lll}
        \multicolumn{1}{c|}{Dataset}         & \multicolumn{3}{|c}{CIFAR100} \\
        \hline
        \multicolumn{1}{c|}{Memory Size M}   &  \multicolumn{1}{|c}{1000} &   \multicolumn{1}{c}{2000}    &   \multicolumn{1}{c}{5000}    \\ 
        \hline\hline
        ER + ours       &   38.5{\scriptsize±0.5} &   45.2{\scriptsize±0.2} &   52.1{\scriptsize±0.5} \\       
        ER + ours (student)      &   37.7{\scriptsize±0.7}       & 44.7{\scriptsize±0.5}      &  51.2{\scriptsize±0.6} \\          
        ER + ours (teacher)     &   37.2{\scriptsize±0.7}       & 43.0{\scriptsize±0.8}       &  49.4{\scriptsize±0.6}  \\          
        ER + ours (student, single view)       &   34.8{\scriptsize±0.6}      & 41.8{\scriptsize±0.6}       &  47.9{\scriptsize±0.4}  \\    
        \hline
    \end{tabular}}
    \label{tab:ablation}
\end{table}

\paragraph{Impact of Multiview Distillation}
As described in the Model Learning section, we employ both augmented and raw images (two views) in our distillation process. In Table~\ref{tab:ablation} we show the performance of \textit{ER + ours} when trained using a single view. Namely, minimizing $\mathcal{L}(X,Y)=\mathcal{L}_{CE}(\hat{X},Y) + \lambda_\alpha KL(\mathcal{T}_{\alpha}(\hat{X}), S(\hat{X}))$. The results indicate that employing this multiview distillation strategy has a significant impact, yielding at least a $2.9\%$ points boost in accuracy.

\section{Discussions}
\label{sec:discussion}
In this section, we analyze the working mechanisms of MKD for OCL.
\vspace{-1em}
\subsection{Choosing $\alpha$}
\begin{figure}[t]
\centering
\includegraphics[width=0.62\linewidth]{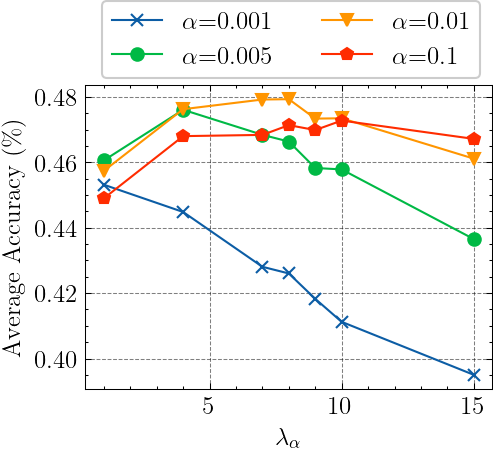}
\caption{Impact of $\lambda_\alpha$ and $\alpha$ on the final performances or \textit{ER} on CIFAR100 M=5k, \textit{clear} setting.}
\label{fig:lambda_alpha1}
\end{figure}

Since $\alpha$ directly influences the teachers' knowledge, it has a significant impact on performances. Finding the best value of $\alpha$ can be done by grid search. Figure \ref{fig:lambda_alpha1} shows the final average accuracy for various values of $(\alpha,\lambda_\alpha)$, in log scale for \textit{ER + Ours} on CIFAR100 M=5K. To avoid computation-intensive grid search, we show in the subsequent section that $\alpha$ can be selected from a broad range, provided the relation between $\alpha$ and $\lambda_\alpha$ is maintained.  
\vspace{-1em}
\subsection{Expressing $\lambda_\alpha$}
\begin{figure}[t]
    \centering
    \includegraphics[width=0.62\linewidth]{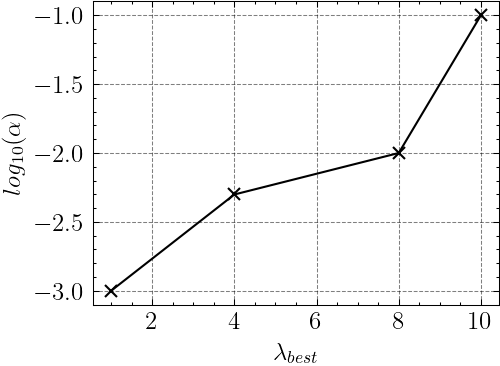}
    \caption{Relation between $\log{\alpha}$ and and the best corresponding $\lambda_\alpha$ value, $\lambda_{best}$. The displayed relation is linear.}
    \label{fig:lambda_alpha2}
\end{figure}

Figure~\ref{fig:lambda_alpha2} illustrates a strong interdependence between $\alpha$ and $\lambda_\alpha$. The optimal value for $\lambda_\alpha$ given $\alpha$ follows the formula $\lambda_\alpha = a*\log_{10}(\alpha) + b$, with $a=9/2$ and $b=29/2$. Notably, lower values of $\alpha$ correspond to lower values of $\lambda$. This correlation arises from the fact that a larger $\alpha$ leads to a teacher closely resembling the student, resulting in a low distillation loss and a higher $\lambda_\alpha$ for compensation.

\vspace{-1em}
\subsection{Reducing Task-Recency Bias}
\begin{figure}[t]
    \centering
    \includegraphics[width=0.9\linewidth]{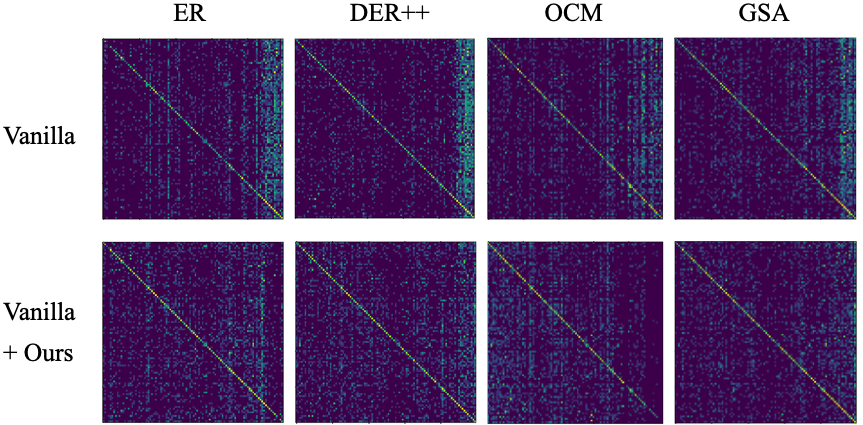}
    \caption{Confusion matrix on the evaluation set at the end of training on CIFAR100 with M=1K for considered baselines. Classes are shown in training order. The top row is the confusion matrices for baselines without the MKD procedure. The bottom row is the confusion matrices when adding MKD.}
    \label{fig:confusion_matrices}
\end{figure}

A common issue in Continual Learning is the task-recency bias~\cite{chrysakis2023online, mai_supervised_2021}. This is the problem of over-predicting the classes belonging to the last task seen. Figure~\ref{fig:confusion_matrices} displays confusion matrices at the end of training for considered baselines, with and without MKD. While most baselines suffer from task-recency bias at the end of training, it can be observed qualitatively that adding MKD reduces this bias by diminishing the amount of last task false positives.
\vspace{-1em}
\subsection{Reducing Last Layer Bias}
Another identified issue when training with Cross Entropy is the presence of bias in the last Fully Connected (FC) layer~\cite{liang_new_2023,ahn_ss-il_2021,mai_supervised_2021, wu_large_2019}. To demonstrate the presence of the last FC bias, one can make use of the Nearest Class Mean (NCM) trick~\cite{mai_supervised_2021} with intermediate representations given by the model. Since we work with memory based approaches, we compare the model's performance using logits with performances obtained by training an NCM classifier using intermediate representation of memory data at the end of training. In other words, we drop the last FC layer and fine-tune with a simple NCM classifier on memory. The NCM trick yields substantial performance improvement in the presence of a pronounced last layer bias, as indicated in Table~\ref{tab:NCM_trick}. Across various baselines, with and without MKD, the NCM trick consistently enhances performances, underscoring the influence of a strong last FC bias. Intriguingly, when our approach is applied to these baselines, leveraging the NCM actually leads to performance degradation. This suggests a neutralization of the last FC layer bias, possibly due to the distillation loss occurring in the logit space, where the last FC layer is tightly constrained.
\begin{table}[t]
    \centering
    \caption{Final Average Accuracy (\%) on CIFAR100 M=1k of several baselines, with and without using the NCM trick. Logits Acc. refers to the accuracy of the model using predicted logits while NCM Acc. refers to NCM accuracy trained on intermediate representations from memory at the end of training.}
    \resizebox{0.62\columnwidth}{!}{
    \begin{tabular}{l|ll}
        Method      & Logits Acc. & NCM Acc. \\
        \hline\hline
        ER~\text{[NeurIPS`19]}              &   23.0{\scriptsize±0.8}    &   29.0{\scriptsize±0.3} ($\uparrow$6.0) \\
        ER + ours                           &   38.5{\scriptsize±0.5}    &   31.5{\scriptsize±0.5} ($\downarrow$7.0)           \\
        DER++~\text{[NeurIPS`20]}           &   22.8{\scriptsize±1.8}    &   26.6{\scriptsize±3.4} ($\uparrow$3.8)           \\
        DER++ + ours                        &   32.1{\scriptsize±0.5}    &   28.7{\scriptsize±1.4} ($\downarrow$3.4)           \\
        ERACE~\text{[ICLR`22]}              &   27.4{\scriptsize±0.6}    &   28.1{\scriptsize±0.7} ($\uparrow$0.7)           \\
        ERACE + ours                        &   37.0{\scriptsize±0.7}    &   34.2{\scriptsize±0.2}  ($\downarrow$2.8)           \\
        GSA~\text{[CVPR`23]}                &   29.7{\scriptsize±1.5}    &   32.6{\scriptsize±1.6} ($\uparrow$2.9)           \\
        GSA + ours                          &   40.1{\scriptsize±1.0}    &   36.2{\scriptsize±0.5} ($\downarrow$3.9)          \\
        \hline
    \end{tabular}}
    \label{tab:NCM_trick}
\end{table}
\vspace{-1em}
\subsection{Reducing Feature Drift}
\begin{figure}[t]
    \centering
    \begin{subfigure}[c]{0.6\linewidth}
    \includegraphics[width=0.97\linewidth]{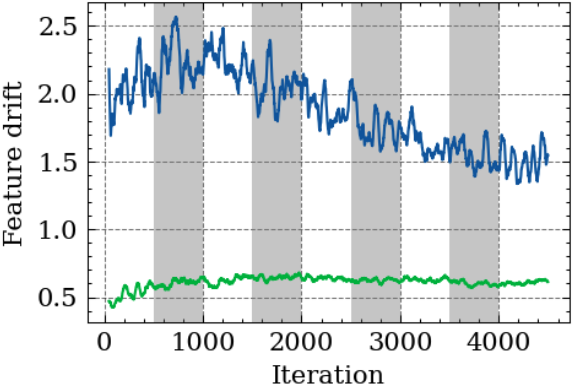}
    \end{subfigure}
    \begin{subfigure}[c]{0.15\linewidth}
        \includegraphics[width=0.97\linewidth]{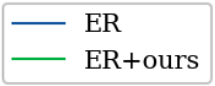}
    \end{subfigure}
    \caption{Feature drift $d_t$ of \textit{ER} and \textit{ER + ours} (MKD) on CIFAR100 M=5k.}
    \label{fig:feature_drift}
\end{figure}

When training in OCL, one potential issue is the feature drift~\cite{caccia_new_2022}. Feature drift occurs when changing tasks causes the representation of old classes to conflict with the representations of new classes, inducing large changes in past representations. Experimentally, we demonstrate that MKD can inherently reduce feature drift. Figure~\ref{fig:feature_drift} shows the feature drift $d_t=\vert\vert f_{\theta_t}(X_{old})-f_{\theta_{t+1}}(X_{old})\vert\vert_2$, where $X_{old}$ are memory images of \textit{old} classes and $f_{\theta_t}$ is the model parameterized by $\theta$ from which we removed the last FC layer. As we can see, using MKD greatly reduces feature drift throughout training. For \textit{ER + ours} (MKD), the feature drift is not only lower but also more stable.

\vspace{-1em}
\subsection{Improving Feature Discrimination}
\begin{figure}[t]
\centering
\begin{subfigure}{0.43\linewidth}
    \includegraphics[width=0.97\linewidth]{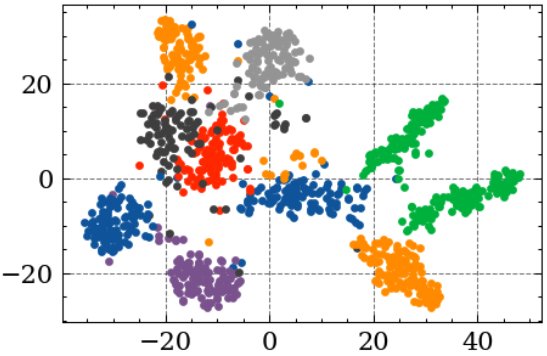}
    \caption{ER}
    \label{fig:first}
\end{subfigure}
\begin{subfigure}{0.43\linewidth}
    \includegraphics[width=0.97\linewidth]{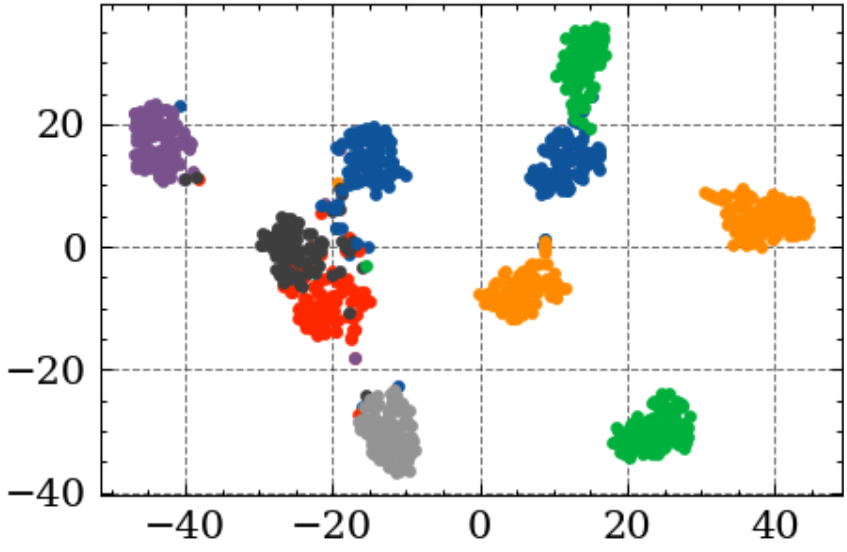}
    \caption{ER + ours (MKD)}
    \label{fig:third}
\end{subfigure}
\begin{subfigure}{0.12\linewidth}
    \includegraphics[width=0.97\linewidth]{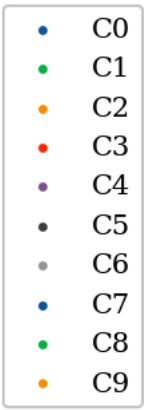}
    \label{fig:sec}
\end{subfigure}
\caption{(a) t-SNE of memory data at the end of training \textit{ER} on CIFAR10, M=1k. (b) t-SNE of memory data at the end of training \textit{ER + ours} (MKD) on CIFAR10, M=1k.}
\label{fig:tsne}
\end{figure}

Feature discrimination is a desirable property of any learning process. Specifically in Continual Learning, it is important to obtain distinctive features at the end of training. In Figure~\ref{fig:tsne}, we present the t-SNE results on memory data at the end of training of \textit{ER} and \textit{ER + ours} (MKD). Clearly, the obtained representation using MKD is significantly more discriminative than the one obtained without MKD. Even though our distillation loss is proposed in the logit space, it can still greatly improve learned feature quality.

\begin{table}[t]
    \centering
    \caption{Backward Transfer (\%) at the end of training on CIFAR100, M=5k and Imagenet100, M=10k for several baselines. Higher is better. Means over 5 runs are displayed.}
    \resizebox{0.62\columnwidth}{!}{
    \begin{tabular}{l|ll}
        Method         &  CIFAR100 & ImageNet100 \\
        \hline\hline
        ER             &     -16.7{\scriptsize±1.2 }& -17.5{\scriptsize±1.5 }\\
        ER + ours      &       +\textbf{8.15{\scriptsize±0.8}}  & -\textbf{1.3{\scriptsize±2.3}} \\
        DER++          &     -27.5{\scriptsize±3.4 } & -18.9{\scriptsize±2.5 }\\
        DER++ + ours   &      -\textbf{10.4{\scriptsize±5.6}}    & -\textbf{14.4{\scriptsize±2.5}} \\
        GSA            &     -4.9{\scriptsize±1.2 }  & -17.0{\scriptsize±1.3 }\\
        GSA + ours     &      -\textbf{2.5{\scriptsize±3.1}}   & -\textbf{15.5{\scriptsize±1.0}} \\
        \hline
    \end{tabular}}
    \label{tab:backward_tf}
\end{table}

\subsection{Improving Backward Transfer}
As the plasticity-stability dilemma is central in CL, a variety of metrics have been designed to adequately measure either plasticity or stability~\cite{mai_online_2021, wang_comprehensive_2023}. We empirically found that leveraging KD in OCL helps retain past information and enhances the model's stability during training. To showcase this effect, we look at the BT of considered baselines, with and without MKD. Table~\ref{tab:backward_tf} shows the BT at the end of training. In every scenario, our method improves BT. Specifically, for \textit{ER}, leveraging MKD can yield a positive BT, implying that the models keep improving on old classes even after a task change. This property is especially important in OCL because the student is unlikely to have fully learned the past task when training on the current task.

\section{Conclusions}
In this paper, we studied the problem of Online Continual Learning from the perspective of Knowledge Distillation. While KD has been widely studied in the context of offline continual learning, it remains under-used in OCL. To understand the current state of KD in OCL, we identified OCL-specific challenges for applying KD: Teacher Quality, Teacher Quantity, and Unknown Task Boundaries. Moreover, we proposed to address these challenges by designing a new distillation procedure based on Momentum Knowledge Distillation. This approach benefits from a powerful plasticity-stability control for OCL and employs an evolving teacher to overcome the previously introduced challenges. We experimentally demonstrated the efficiency of our approach and achieved more than $10\%$ points improvement over state-of-the-art methods on several datasets. Additionally, we provided insightful explanations on how using MKD can help solve multiple OCL known issues: task-recency bias, last layer bias, feature drift, feature discrimination, and backward transfer. Our approach is architecture-independent and computationally efficient. In conclusion, we have shed new light on distillation for OCL and advocate for its efficiency and its potential as a central component for addressing OCL.

\section*{Acknowledgments}
This work has received support from Agence Nationale de la Recherche (ANR) for the project APY, with reference ANR-20-CE38-0011-02 and was granted access to the HPC resources of IDRIS under the allocation 2022-AD011012603 made by GENCI. This work benefited from an international mobility grant from Paris Est Sup which enabled the collaboration between the Gustave Eiffel University and the University of Tokyo.

\section*{Impact Statement}
This paper presents work whose goal is to advance the field of Machine Learning. There are many potential societal consequences of our work, none which we feel must be specifically highlighted here.

\bibliography{references}
\bibliographystyle{icml2024}

\newpage
\appendix
\onecolumn
\section{Additional Experiments}

\subsection{Task-Recency Bias}

\begin{figure*}[ht]
    \centering
    \includegraphics[width=1.0\linewidth]{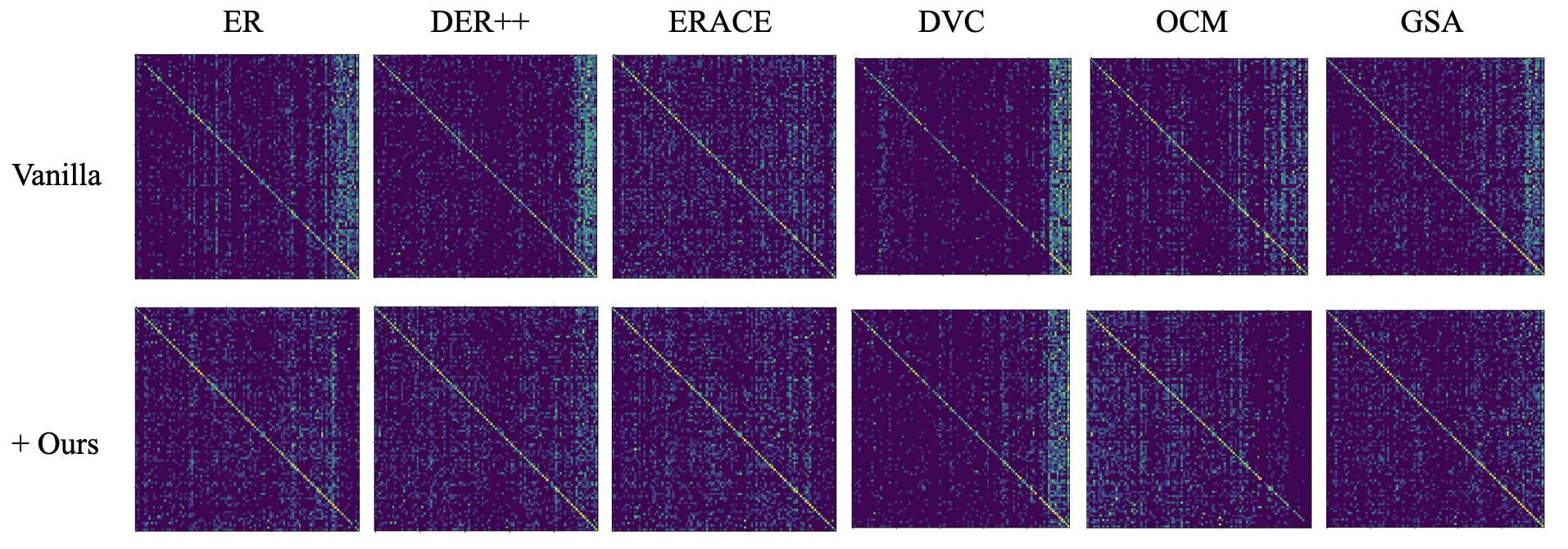}
    \caption{Confusion matrix on the evaluation set at the end of training on CIFAR100 with M=1K for considered baselines. Classes are shown in the same order of those during training such that left columns of confusion matrices correspond to first classes seen during training. The top row presents the confusion matrices for baselines without the MKD procedure. The bottom row is the confusion matrices when adding MKD.}
    \label{fig:supp_confusion_matrices}
\end{figure*}

In the main paper, we discussed how our approach addresses the task-recency bias in OCL for only a limited number of methods due to space constraints. In Figure~\ref{fig:supp_confusion_matrices}, we share confusion matrices for every considered method from the main paper.

\subsection{Last layer Bias}
In Table~\ref{tab:all_ncm_trick} we share extra experiments regarding the impact of the NCM trick. Specifically, OCM and DVC are not included in the main paper.
\begin{table}[ht]
    \centering
    \resizebox{0.5\columnwidth}{!}{
    \begin{tabular}{l|ll}
        Method      & Logits Acc. & NCM Acc. \\
        \hline\hline
        ER~\text{[NeurIPS`19]}              &   23.0{\scriptsize±0.8}    &   29.0{\scriptsize±0.3} ($\uparrow$6.0) \\
        ER + ours                           &   38.5{\scriptsize±0.5}    &   31.5{\scriptsize±0.5} ($\downarrow$7.0)           \\
        DER++~\text{[NeurIPS`20]}           &   22.8{\scriptsize±1.8}    &   26.6{\scriptsize±3.4} ($\uparrow$3.8)           \\
        DER++ + ours                        &   32.1{\scriptsize±0.5}    &   28.7{\scriptsize±1.4} ($\downarrow$3.4)           \\
        DVC~\text{[CVPR`22]}           &   19.5{\scriptsize±2.1}    &   21.1{\scriptsize±1.5} ($\uparrow$1.6)           \\
        DVC + ours                        &   27.0{\scriptsize±2.0}    &   27.4{\scriptsize±1.7} ($\uparrow$0.4)           \\
        ERACE~\text{[ICLR`22]}              &   27.4{\scriptsize±0.6}    &   28.1{\scriptsize±0.7} ($\uparrow$0.7)           \\
        ERACE + ours                        &   37.0{\scriptsize±0.7}    &   34.2{\scriptsize±0.2}  ($\downarrow$2.8)           \\
        OCM~\text{[ICML`22]}                &   29.1{\scriptsize±1.4}    &   29.3{\scriptsize±1.2} ($\uparrow$0.2)           \\
        OCM + ours                          &   37.1{\scriptsize±0.7}    &   31.6{\scriptsize±0.2} ($\downarrow$5.5)          \\
        GSA~\text{[CVPR`23]}                &   29.7{\scriptsize±1.5}    &   32.6{\scriptsize±1.6} ($\uparrow$2.9)           \\
        GSA + ours                          &   40.1{\scriptsize±1.0}    &   36.2{\scriptsize±0.5} ($\downarrow$3.9)          \\
        \hline
        
    \end{tabular}}
    \caption{Final Average Accuracy (\%) on CIFAR100 M=1k of various baselines, with and without using the NCM trick. Logits Acc. refers to the accuracy of the model using predicted logits while NCM Acc. refers to NCM accuracy trained on intermediate representations from memory at the end of training.}
    \label{tab:all_ncm_trick}
\end{table}

\subsection{Feature Drift}
\begin{figure*}[t]
\centering
\begin{subfigure}[c]{0.3\textwidth}
    \includegraphics[width=0.9\linewidth]{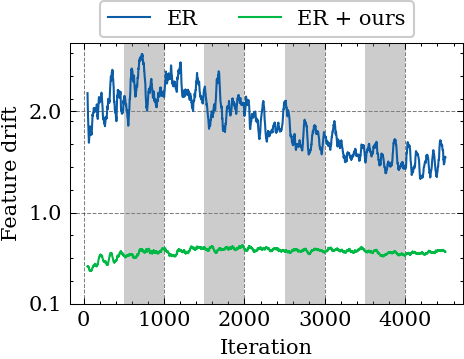}
\end{subfigure}
\begin{subfigure}[c]{0.3\textwidth}
    \includegraphics[width=0.93\linewidth]{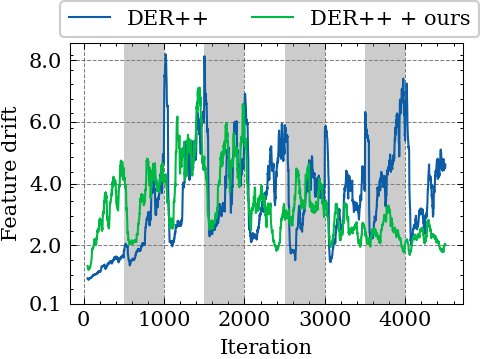}
\end{subfigure}
\begin{subfigure}[c]{0.3\linewidth}
    \includegraphics[width=0.9\linewidth]{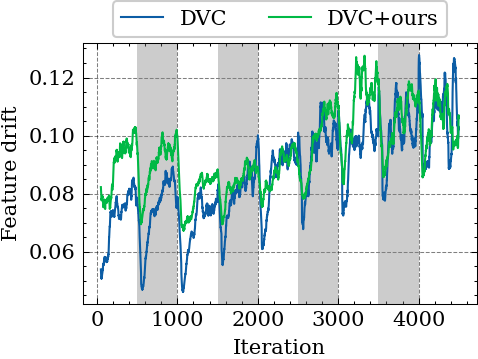}
\end{subfigure}
\begin{subfigure}[c]{0.3\linewidth}
    \includegraphics[width=0.9\linewidth]{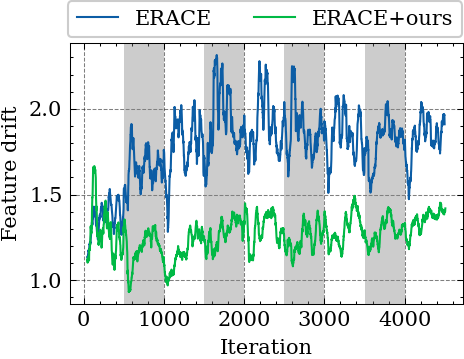}
\end{subfigure}
\begin{subfigure}[c]{0.3\linewidth}
    \includegraphics[width=0.9\linewidth]{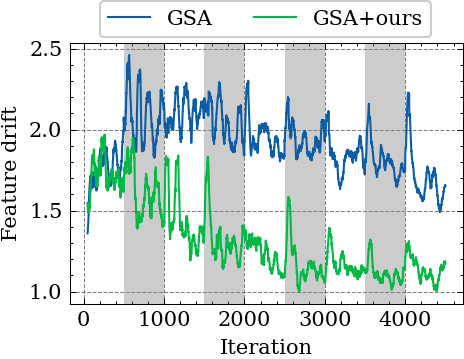}
\end{subfigure}
\begin{subfigure}[c]{0.3\linewidth}
    \includegraphics[width=0.9\linewidth]{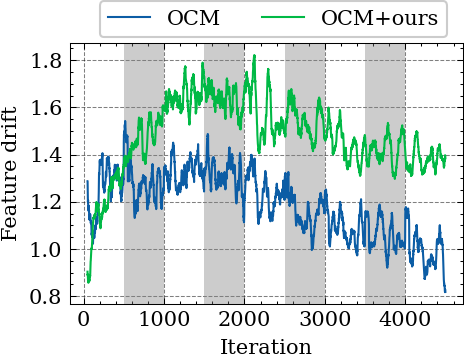}
\end{subfigure}
\caption{Feature drift $d_t$ of \textit{ER}, \textit{DER++}, \textit{DVC}, \textit{ERACE}, \textit{GSA}, \textit{OCM} and their MKD adaptations on CIFAR100, M=5k.}
\label{fig:supp_feat_drift}
\end{figure*}

We show additional experiments concerning the impact of MKD on feature drift on Figure~\ref{fig:supp_feat_drift}. It can be observed that for \textit{GSA} and \textit{ERACE}, introducing MKD can greatly help in reducing feature drift. However, this phenomenon is not as pronounced with \textit{DVC} and \textit{DER++}. Since \textit{DVC} encourages representations to be augmentation-invariant, it is expected to observe more stability against feature drift with \textit{DVC}. Notably, the drift values of \textit{DVC} and \textit{DVC + ours} are considerably lower than any other considered method. 
Additionaly, we observe the opposite effect for OCM, which also incorporate feature stability by leveraging a contrastive objective~\cite{guo_online_2022}. Even though MKD cannot reduce feature drift for OCM, experimental results still demonstrate a significant improvement in performances.

\subsection{Feature Discrimination}
\begin{figure*}[t]
\centering
\begin{subfigure}[c]{0.24\textwidth}
    \includegraphics[width=0.85\linewidth]{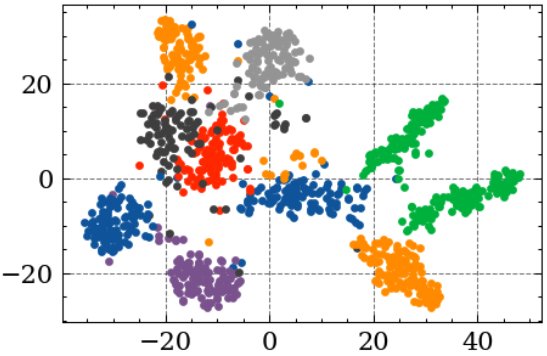}
    \caption{ER}
\end{subfigure}
\begin{subfigure}[c]{0.24\textwidth}
    \includegraphics[width=0.85\linewidth]{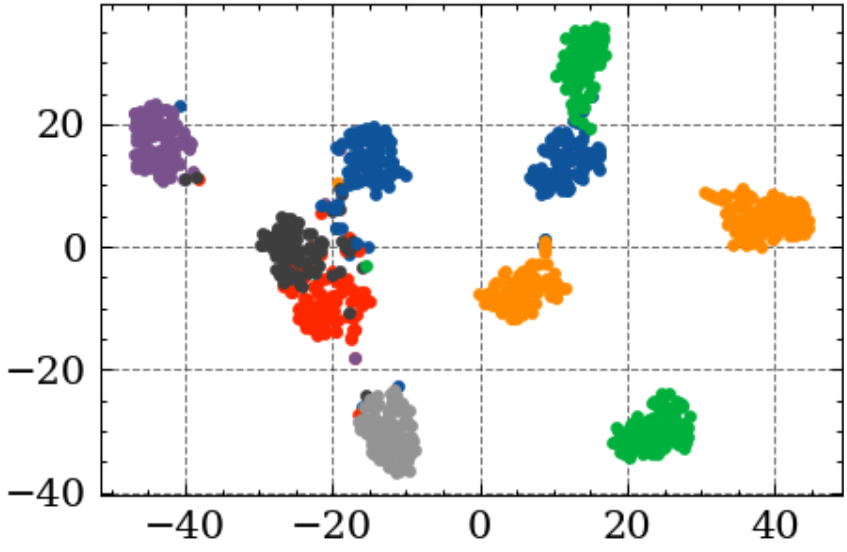}
       \caption{ER + ours}
\end{subfigure}
\begin{subfigure}[c]{0.24\linewidth}
    \includegraphics[width=0.85\linewidth]{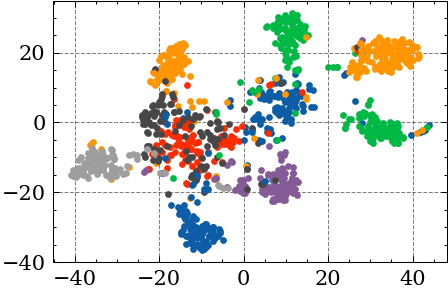}
    \caption{DER++}
\end{subfigure}
\begin{subfigure}[c]{0.24\linewidth}
    \includegraphics[width=0.85\linewidth]{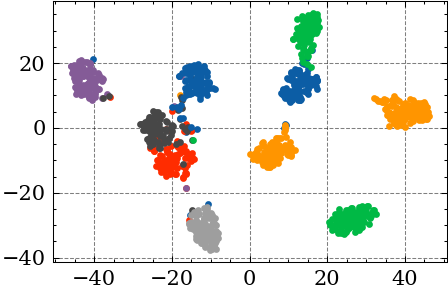}
    \caption{DER++ + ours}
\end{subfigure}

\begin{subfigure}[c]{0.24\linewidth}
    \includegraphics[width=0.85\linewidth]{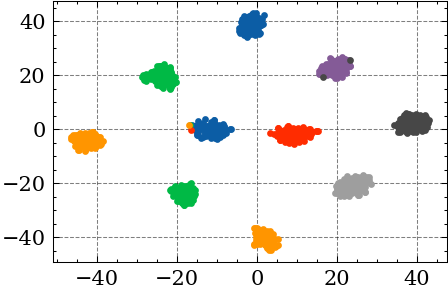}
    \caption{DVC}
\end{subfigure}
\begin{subfigure}[c]{0.24\linewidth}
    \includegraphics[width=0.85\linewidth]{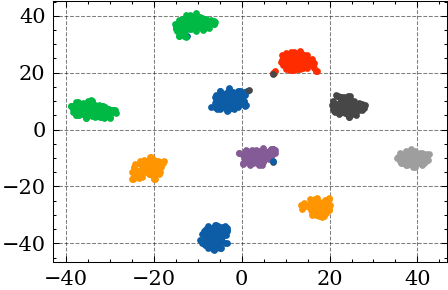}
    \caption{DVC + ours}
\end{subfigure}
\begin{subfigure}[c]{0.24\linewidth}
    \includegraphics[width=0.85\linewidth]{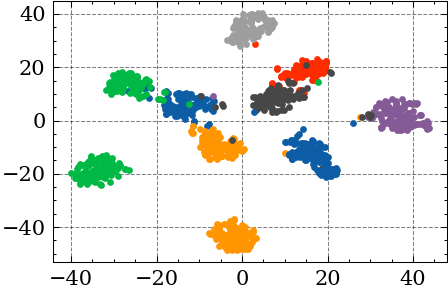}
    \caption{ERACE}
\end{subfigure}
\begin{subfigure}[c]{0.24\linewidth}
    \includegraphics[width=0.85\linewidth]{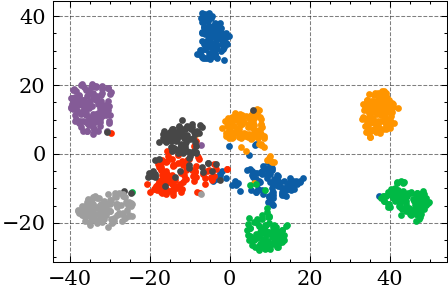}
    \caption{ERACE + ours} 
\end{subfigure}

\begin{subfigure}[c]{0.24\linewidth}
    \includegraphics[width=0.85\linewidth]{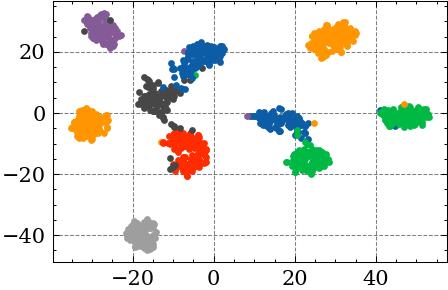}
    \caption{GSA}
\end{subfigure}
\begin{subfigure}[c]{0.24\linewidth}
    \includegraphics[width=0.85\linewidth]{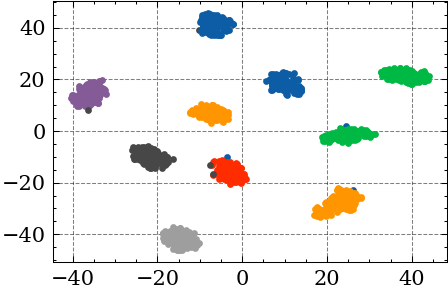}
    \caption{GSA + ours}
\end{subfigure}
\begin{subfigure}[c]{0.24\linewidth}
    \includegraphics[width=0.85\linewidth]{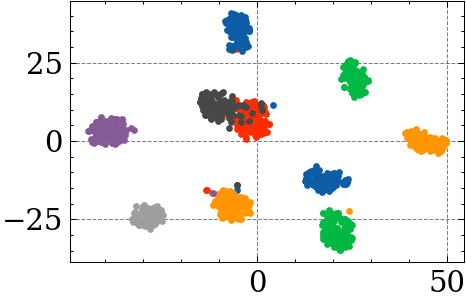}
   \caption{OCM}
\end{subfigure}
\begin{subfigure}[c]{0.24\linewidth}
    \includegraphics[width=0.85\linewidth]{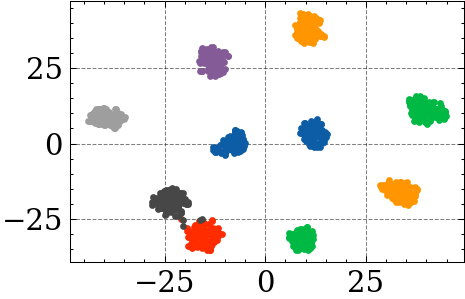}
    \caption{OCM + ours}
\end{subfigure}
\caption{t-SNE visualization of \textit{ER}, \textit{DER++}, \textit{DVC}, \textit{ERACE}, \textit{GSA}, \textit{OCM} and their MKD adaptations on CIFAR100, M=5k.}
\label{fig:all_tsne}
\end{figure*}

To showcase the impact of MKD on feature discrimination, we presented t-SNE results on memory data at the end of training for \textit{ER} and \textit{ER + ours}. In Figure~\ref{fig:all_tsne} we present additional t-SNE experiments for remaining baselines. We used a perplexity of $30$ for these experiments.

\subsection{Backward Transfer}
In Table~\ref{tab:backward_tf_app} we present additional experiments concerning the impact of MKD on Backward Transfer (BT). Specifically, OCM and DVC are not included in the main paper because of the limited space.
\begin{table}[t]
    \centering
    \resizebox{0.35\columnwidth}{!}{
    \begin{tabular}{l|ll}
        Method         &  CIFAR100 & ImageNet100 \\
        \hline\hline
        ER             &     -16.7{\scriptsize±1.2 }                &           -17.5{\scriptsize±1.5 }\\
        ER + ours      &       +\textbf{8.15{\scriptsize±0.8}}      &           -\textbf{1.3{\scriptsize±2.3}} \\
        DER++          &     -27.5{\scriptsize±3.4 }                &           -18.9{\scriptsize±2.5 }\\
        DER++ + ours   &      -\textbf{10.4{\scriptsize±5.6}}       &           -\textbf{14.4{\scriptsize±2.5}} \\
        ERACE          &        -3.0{\scriptsize±1.6}              &           -6.5{\scriptsize ±1.3}\\
        ERACE + ours   &      +\textbf{8.6{\scriptsize±1.6}}       &           -\textbf{5.6{\scriptsize±1.4}} \\
        DVC             &     -32.5{\scriptsize ±4.7}             &           -34.3{\scriptsize ±5.9} \\
        DVC + ours      &     -\textbf{20.9{\scriptsize ±6.6}}      &       -\textbf{31.5{\scriptsize ±0.8}}    \\
        OCM             &     -5.0±{\scriptsize 3.0}              &           +1.1±{\scriptsize 1.6}\\
        OCM + ours      &       +\textbf{14{\scriptsize±1.8}}       &       +\textbf{3.9{\scriptsize ±0.8}}     \\
        GSA            &     -4.9{\scriptsize±1.2 }                 &           -17.0{\scriptsize±1.3 }\\
        GSA + ours     &      -\textbf{2.5{\scriptsize±3.1}}        &           -\textbf{15.5{\scriptsize±1.0}} \\
        \hline
    \end{tabular}}
    \caption{Backward Transfer (\%) at the end of training on CIFAR100, M=5k and Imagenet100, M=10k for various baselines. Higher is better. Mean and standard deviations over 5 runs are displayed.}
    \label{tab:backward_tf_app}
\end{table}

\section{Experimental Details}
\subsection{Datasets}
We use variations of standard image classification datasets~\cite{krizhevsky_learning_2009,le_tiny_2015,deng2009imagenet}. The original datasets are split into several tasks of non-overlapping classes. Specifically, we experimented on CIFAR10, CIFAR100, Tiny ImageNet, and ImageNet-100.\\
\textbf{CIFAR10} contains 50,000 32x32 train images and 10,000 test images and is split into 5 tasks, each containing 2 classes, for a total of 10 distinct classes.\\
\textbf{CIFAR100} contains 50,000 32x32 train images and 10,000 test images and is split into 10 tasks, each contains 10 classes, for a total of 100 distinct classes.\\
\textbf{Tiny ImageNet} is a subset of the ILSVRC-2012 classification dataset and contains 100,000 64x64 train images as well as 10,000 test images and is split into 20 tasks, each containing 10 classes,  for a total of 200 distinct classes.\\
\textbf{ImageNet-100} is another subset of ILSVRC-2012 containing only the first 100 classes with 1,300 224x224 images per class for training and 50 for testing.

\subsection{Data Augmentation}
Several methods have demonstrated improved performance through the use of simple augmentations rather than more intricate ones. To ensure optimal performance comparison among the various methods, we employed two distinct augmentation strategies: the \textit{partial} and the \textit{full} strategies.

\paragraph{\textit{Partial} Augmentation Strategy.} The \textit{partial} augmentation strategy comprises only a subset of the augmentations utilized in the \textit{full} strategy. Specifically, it involves a sequence of random cropping and random horizontal flipping, both with a probability $p$ of 0.5.

\paragraph{\textit{Full} Augmentation Strategy.} The \textit{full} augmentation strategy encompasses a wider array of augmentations. It involves a sequence of random cropping, horizontal flipping, color jitter, and random grayscale transformations. The parameters for color jitter are set to $(0.4, 0.4, 0.4, 0.1)$ with a probability $p$ of 0.8. The application probability for random grayscale is set at 0.2.

These strategies have also been chosen during the hyper-parameter search.

\subsection{Task boundaries inference}
For experimenting on the \textit{blurry} setting with OCM~\cite{guo_online_2022}, it is necessary to infer the task change. Inferring task change in this setup can be cumbersome and grandly impact performances. For simplicity, we detect task change by applying two simple rules. We consider the task has changed if:
\begin{itemize}
    \item A new class (never seen by the model) appears in the stream;
    \item The last task change appeared at least $100$ iterations previous to the current one.
\end{itemize}

\subsection{Hyper-parameters table}
Different hyper-parameters values used in grid search for considered methods are reported in Table~\ref{tab:all_hp}. This grid search has been conducted on CIFAR100, M=5k. Note that we used parameters from the original paper for OCM~\cite{guo_online_2022} due to computational constraints.
\begin{table}[!ht]
    \setlength{\tabcolsep}{4pt}
    \centering
    \resizebox{0.6\columnwidth}{!}{\begin{tabular}{c|cc}
    Method & Parameter & Values \\
    \midrule
    \multirow{5}*{ER}    & optim         & [SGD, Adam] \\
                        & weight decay  & [0, 1e-4] \\
                        & lr            & [0.0001, 0.001, 0.01, 0.1] \\
                        & momentum      & [0, 0.9] \\
                        & aug. strat.   & [full, partial] \\
     \midrule
    \multirow{5}*{ER-ACE}   & optim         & [SGD, Adam] \\
                        & weight decay  & [0, 1e-4] \\
                        & lr            & [0.0001, 0.001, 0.01, 0.1] \\
                        & momentum      & [0, 0.9] \\
                        & aug. strat.   & [full, partial] \\
    \midrule
    \multirow{7}*{DER++}    & optim         & [SGD, Adam] \\
                        & weight decay  & [0, 1e-4] \\
                        & lr            & [0.0001, 0.001, 0.01, 0.03] \\
                        & momentum      & [0, 0.9] \\
                        & aug. strat.   & [full, partial] \\
                        & alpha         & [0.1, 0.2, 0.5, 1.0] \\
                        & beta          & [0.5, 1.0] \\
    \midrule
    \multirow{5}*{DVC}   & optim         & [SGD, Adam] \\
                        & weight decay  & [0, 1e-4] \\
                        & lr            & [0.0001, 0.001, 0.01, 0.1] \\
                        & momentum      & [0, 0.9] \\
                        & aug. strat.   & [full, partial] \\
    \midrule
    \multirow{5}*{GSA}  & optim         & [SGD, Adam] \\
                        & weight decay  & [0] \\
                        & lr            & [0.0001, 0.0005, 0.01, 0.05, 0.01] \\
                        & momentum      & [0] \\
                        & aug. strat.   & [full, partial] \\
    \midrule
    \multirow{5}*{ER+SDP}  & optim         & [Adam] \\
                        & weight decay  & [0] \\
                        & lr            & [0.0003] \\
                        & momentum      & [0] \\
                        & $\mu $          & [10, 100, 1000, 10000] \\
                        & $c^2$          & [0.5, 0.75, 0.9] \\
    \midrule
    \multirow{5}*{PCR}   & optim         & [Adam] \\
                        & weight decay  & [0] \\
                        & lr            & [0.0005] \\
                        & momentum      & [0.9] \\
                        & aug. strat.   & [full] \\
    \midrule
    \multirow{6}*{ER+Temp. Ens.}   & optim         & [Adam] \\
                        & weight decay  & [0] \\
                        & lr            & [0.0005] \\
                        & momentum      & [0.9] \\
                        & aug. strat.   & [full] \\
                        & $\alpha$      & [0.01, 0.05, 0.1, 0.5, 0.9, 0.95, 0.99] \\
                    \end{tabular}}
\caption{Hyper-parameters tested for every method on CIFAR100, M=5k, 10 tasks.\label{tab:all_hp}}
\end{table}

\subsection{Hardware and computation}

\begin{figure*}[t]
    \centering
    \includegraphics[width=0.5\linewidth]{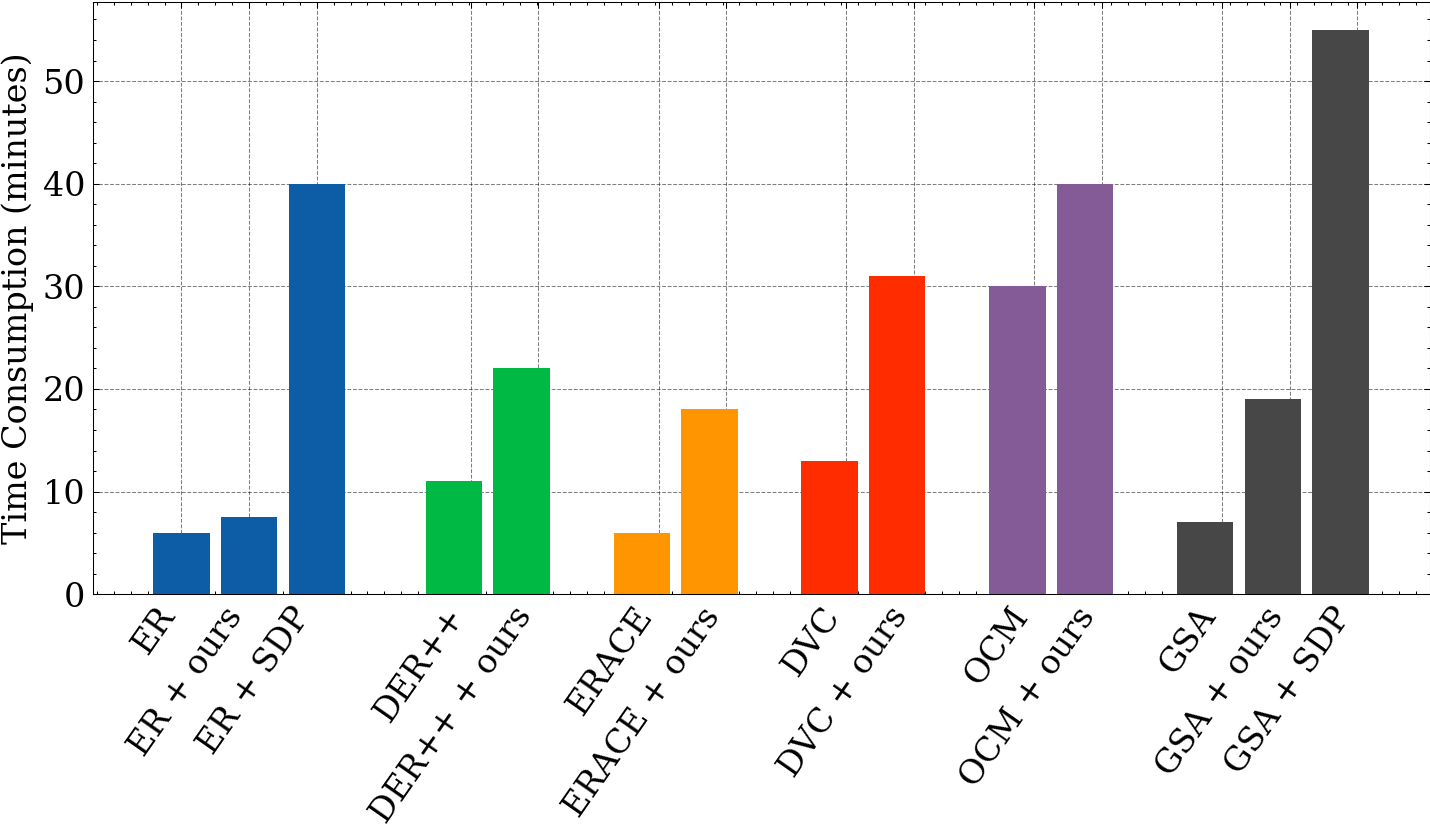}
    \caption{Time consumption (minutes) of compared methods when training on CIFAR100, M=5k with V100 GPUs.}
    \label{fig:time_consumption}
\end{figure*}

For the compared methods, we trained on RTX A5000 and V100 GPUs. Figure~\ref{fig:time_consumption} references the training time of each method on CIFAR100 M=5k.

\end{document}